\documentclass[10pt,twocolumn,letterpaper]{article}

\usepackage{wacv}              

\usepackage{graphicx}
\usepackage{adjustbox}
\usepackage{rotating} 
\usepackage{booktabs}
\usepackage{multirow}
\usepackage[table]{xcolor} 
\usepackage{pifont}        
\usepackage{longtable}
\usepackage{algorithm}
\usepackage{algpseudocode}
\usepackage{pgffor}
\usepackage[most]{tcolorbox}
\usepackage[T1]{fontenc}
\usepackage{newtxtext}
\usepackage[normalem]{ulem} 
\definecolor{myblue}{HTML}{3B6BA5}
\definecolor{mygreen}{HTML}{5B9759}
\definecolor{myorange}{HTML}{B1752D}

\newtcolorbox{prompt}{ 
    title=Prompt, 
    enhanced, 
    colback=myblue!5!white,       
    colframe=myblue!90!white,     
    colbacktitle=myblue!90!white, 
    coltitle=white,               
    fonttitle=\bfseries,          
    arc=0mm,                      
    boxrule=1pt,                  
    left=6pt, right=6pt, top=6pt, bottom=6pt, 
    toptitle=4pt,                 
    bottomtitle=4pt,              
    lefttitle=6pt,                
    righttitle=6pt                
}

\newcommand{\yannis}[1]{{#1}}

\newcommand{\mypar}[1]{\vspace{0mm}\noindent\textbf{#1}}

\newcommand{\numscenes}{$27{,}010$ }
\newcommand{\numannotation}{$27$M }
\newcommand{\modelicon}[1]{\raisebox{-0.2\height}{\includegraphics[height=1.1em]{imgs/icons/#1}}\,}

\definecolor{wacvblue}{rgb}{0.21,0.49,0.74}
\usepackage[pagebackref,breaklinks,colorlinks,allcolors=wacvblue]{hyperref}

\def\wacvPaperID{274} 
\def\confName{WACV}
\def\confYear{2027}

\title{{HiddenObjects}: Scalable Spatial Prior Extraction for Object Placement from Diffusion Models}

\author{
    Marco Schouten$^{1,3}$ \quad 
    Ioannis Siglidis$^{2,3}$ \quad 
    Serge Belongie$^{2,3}$ \quad 
    Dim P. Papadopoulos$^{1,3}$ \\
    $^{1}$Technical University of Denmark \quad 
    $^{2}$University of Copenhagen \quad 
    $^{3}$Pioneer Centre for AI \\
    {\large\url{https://hidden-objects.github.io/}}
}

\definecolor{vlmblue}{HTML}{2F5D8A}
\definecolor{vlmbluebg}{HTML}{F2F6FA}
\definecolor{vlmgray}{HTML}{59636E}
\definecolor{vlmgraybg}{HTML}{F6F7F8}

\newtcolorbox{vlmprompt}[1]{%
  enhanced,
  breakable,
  colback=vlmbluebg,
  colframe=vlmblue,
  coltitle=white,
  colbacktitle=vlmblue,
  title={#1},
  fonttitle=\bfseries\small,
  boxrule=0.6pt,
  arc=1.5pt,
  left=6pt,
  right=6pt,
  top=5pt,
  bottom=5pt,
  before skip=6pt,
  after skip=8pt
}

\newcommand{\promptrole}[1]{\textcolor{vlmblue}{\bfseries #1}}
\newcommand{\promptvar}[1]{\textcolor{vlmgray}{\texttt{[#1]}}}

\begin{document}
\twocolumn[{%
 \renewcommand\twocolumn[1][]{#1}
 \maketitle

 \centering

\begin{minipage}[t]{0.615\textwidth}
    \centering
    \includegraphics[height=6cm,keepaspectratio]{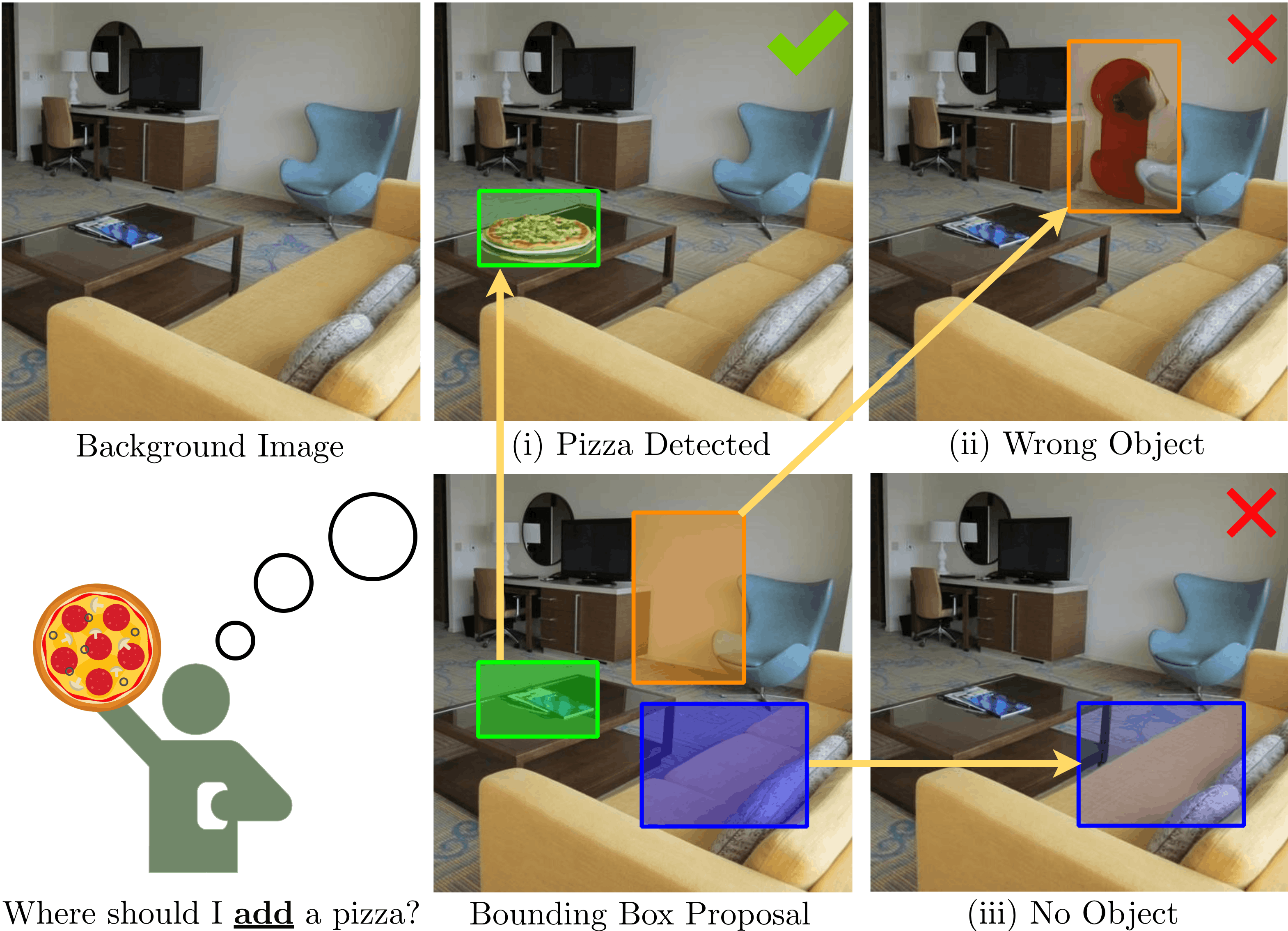}
    \par\vspace{0.5em}\small (a) Spatial Prior Extraction
\end{minipage}\hspace{0.008\textwidth}%
\begin{minipage}[t]{0.377\textwidth}
    \centering
    \includegraphics[height=6cm,keepaspectratio]{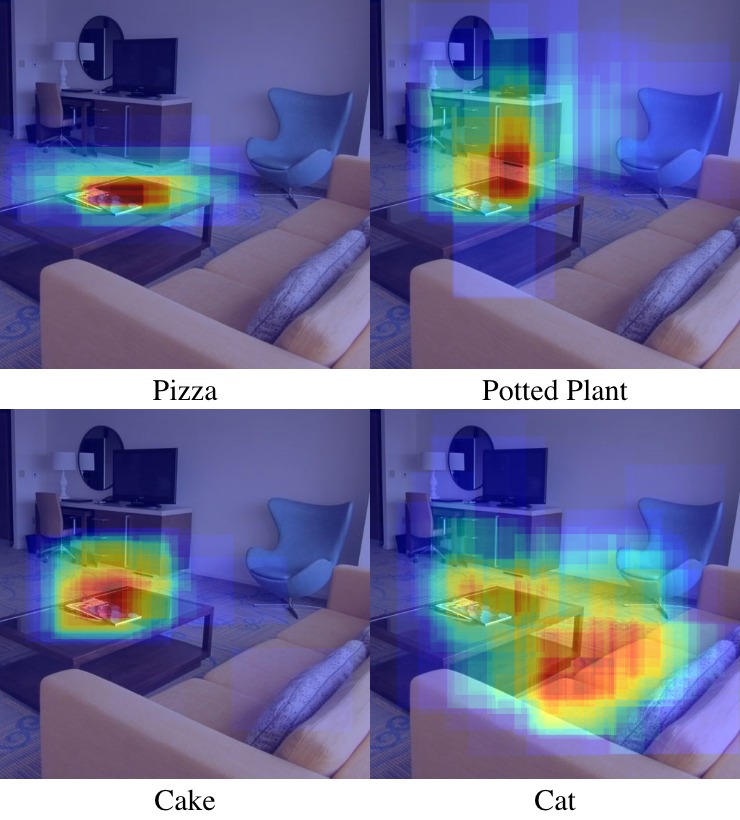}
    \par\vspace{0.5em}\small (b) Multi-class Spatial Priors
\end{minipage}
\par
\vspace{1em}
\captionsetup{hypcap=false} 
\captionof{figure}{\textbf{Spatial Priors for Object Placements}. (a) We score candidate bounding boxes for object insertion using a diffusion inpainting pipeline. In this example, we consider three insertion locations by inpainting and evaluate each location based on three possible outcomes: (i) {\textcolor{green!50!black}{``Pizza Detected''}}, (ii) {\textcolor{orange!80!black}{``Wrong Object''}}, or (iii) {\textcolor{blue!80!black}{``No Object''}}. Plausible placements (i.e., (i)) contribute to the prior. (b) Aggregating predictions across many scenes and object categories produces heatmaps reflecting common compositional patterns in natural scenes.}
\label{fig:teaser}
\vspace{0.5em}
}]

\begin{abstract}
  We learn explicit, class-conditioned spatial priors for object placement in natural scenes by
  distilling placement knowledge from text-conditioned diffusion models. Prior methods rely on sparse manual annotations
  or on inpainting object-removal pipelines, whose
artifacts promote shortcut learning. To address these limitations, we propose a scalable pipeline that evaluates dense candidate placements by
  inserting objects into clean backgrounds. This produces
  \textsc{HiddenObjects}, a dataset of 27M bounding-box annotations (2M positives) across 27K diverse scenes and object categories. Human evaluation prefers our image editing composites over those guided by sparse
  annotations in 65.6\% of comparisons. We distill the priors into a lightweight placement
  model that achieves the best accuracy among dedicated baselines and
  zero-shot vision-language models.
  \end{abstract}    
\providecommand{\cut}[1]{\sout{#1}}
\providecommand{\new}[1]{\textcolor{blue}{#1}}

\section{Introduction}

Should a pizza ever appear on the floor? As products of human priors, natural scene images reflect strong spatial regularities: pizzas rest on tables, books lie on shelves, and clocks hang on walls. These contextual spatial priors have long been implicitly exploited to improve scene understanding~\cite{torralba2003context,rabinovich2007objects,galleguillos2008object,doersch2015unsupervised}. While these priors underlie most visual datasets, they have not been explicitly modeled at scale. Modeling these spatial priors enables applications, such as realistic image editing~\cite{schouten2025poempreciseobjectlevelediting,chen2026referring} and outlier detection~\cite{BoukercheZA20}.

\begin{figure}[t]
  \centering
  \includegraphics[width=\linewidth]{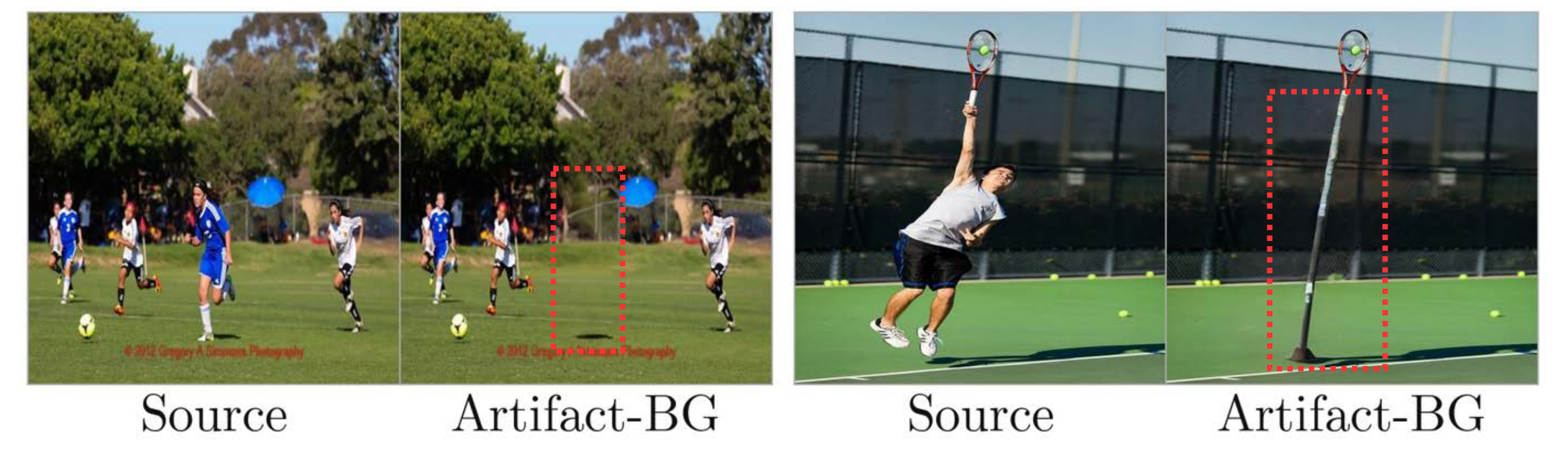}
  \caption{\textbf{Background artifacts in PIPE~\cite{wasserman2024paint}.} Object removal pipelines are effective, but leave visible traces that facilitate shortcut learning, e.g., (left) blurry area or (right) larger artifacts.}
  \label{fig:artefact}
\end{figure}

Recent object placement work extracts spatial priors either by removing existing objects from images via inpainting~\cite{wasserman2024paint,parihar2024text2place,winter2025objectmate} or by manually annotating placements for new objects~\cite{liu2021opa,li2023dreamedit}. While the first approach can produce priors automatically, the removed objects can contain background artifacts (see Fig.~\ref{fig:artefact}) that facilitate shortcut learning. The second approach avoids background artifacts, but as manual annotation is not scalable, it becomes a trade-off between low-dataset diversity (as not a lot of images can be annotated) and unconfident annotations (as images can only be annotated by few people). In both cases, datasets contain sparse object placements (i.e., one location per instance) and likelihoods associated with these proposals.

Instead, our main observation is that spatial priors are closely related to image completion: given a scene image, predicting plausible object placements is essentially a matter of filling in missing content. Early non-parametric approaches showed that plausible completions can be retrieved from the context of millions of images~\cite{hays2007scene}. Interestingly, even early GAN-based image editing tools revealed that generative models implicitly learn contextual constraints. They understand where objects belong and resist placing them in implausible regions (e.g., refusing to draw doors in the sky)~\cite{bau2018gan,bau2020units,papadopoulos2019make} enabled by training on 
billions of images~\cite{schuhmann2022laion}. Unlike previous approaches, diffusion models offer fine-grained controllability, allowing users to add objects across all plausible locations of an input scene.

 Motivated by these abilities, we propose a framework that extracts scalable, dense, and explicit spatial priors directly from diffusion models without manual bounding-box annotations.
Practically, our key observation is that while diffusion models can inpaint objects in semantically plausible locations (Fig.~\ref{fig:teaser}.i), they can also produce wrong inpaintings (Fig.~\ref{fig:teaser}.ii) or refuse inpainting altogether (Fig.~\ref{fig:teaser}.iii). Using state-of-the-art object detectors~\cite{ren2024grounded} we can verify the validity of an insertion. Then, using an image-based reward model~\cite{xu2023imagereward}, we can rank preferences across positive placements. Exhaustive testing of multiple insertions using a sliding window can densely capture all plausible object locations within a scene, finally producing explicit, class-conditioned spatial priors (Fig.~\ref{fig:teaser}b).

Using our pipeline, we construct a large-scale dataset of 27M object placement annotations (2M positive) in 27K scenes from Places365~\cite{zhou2017places}. Moreover, we distill these priors into a transformer-based model that predicts bounding boxes for new scenes efficiently, enabling fast inference for downstream applications. Our experimental results reveal three key findings. 
First, on the downstream task of object insertion, our automatically generated annotations yield significantly higher image editing quality than sparse manual annotations preferred in 65.6\% of comparisons (3.90 vs. 2.68 average ImgEdit-Judge score~\cite{ye2025imgedit}), indicating that sparse single-point annotations are weaker targets for diffusion-based inpainting.
Second, the extracted priors are consistent with common compositional patterns while exhibiting greater spatial diversity than object-centric datasets. 
Third, our distilled model achieves the best overall placement accuracy (mAP) among dedicated baselines and zero-shot VLMs, confirming that dense, reward-weighted spatial priors are essential for learning robust placement distributions.

Our contributions are threefold:
\noindent\textbf{(1)} \yannis{We} present a scalable framework to extract multi-class spatial priors from diffusion models for input images.
\textbf{(2)} \yannis{We} construct \textsc{HiddenObjects}, a dataset of 27M densely evaluated object placements in 27K images.
\textbf{(3)} 
\yannis{We} distill these priors into a lightweight transformer that achieves a 230,000$\times$ speedup and 300$\times$ memory reduction over the full exhaustive extraction pipeline, enabling fast object placement proposals for downstream applications.

\section{Related Work}

\begin{table*}[t]
\centering
\caption{\textbf{Comparison of Object Placement Datasets.}  Manually annotated datasets are limited in scale, while automated approaches introduce artifacts via inpainting. \textsc{HiddenObjects} provides scalable, dense bounding-box annotations on artifact-free backgrounds.}
\label{tab:dataset_stats_dense}
\resizebox{\textwidth}{!}{%
\begin{tabular}{lccccccc}
\toprule
\multirow{2}{*}{\textbf{Dataset}}
& \textbf{Annotations}
& \textbf{Positive Boxes}
& \textbf{Negative Boxes}
& \textbf{Real}
& \textbf{Dense}
& \multirow{2}{*}{\textbf{Scalable}}
& \textbf{Artifact-free} \\
& \textbf{Boxes}
& \textbf{(per instance)}
& \textbf{(per instance)}
& \textbf{Images}
& \textbf{Priors}
&
& \textbf{Background} \\
\midrule
DreamEditBench~\cite{li2023dreamedit}
& 220 & 1 & 0
& \textcolor{green!70!black}{\ding{51}}
& \textcolor{red}{\ding{55}}
& \textcolor{red}{\ding{55}}
& \textcolor{green!70!black}{\ding{51}} \\
MureCom~\cite{lu2023dreamcom}
& 640 & 1 & 0
& \textcolor{green!70!black}{\ding{51}}
& \textcolor{red}{\ding{55}}
& \textcolor{red}{\ding{55}}
& \textcolor{green!70!black}{\ding{51}} \\
OPA~\cite{liu2021opa}
& 0.07M & 1.9 & 3.8
& \textcolor{green!70!black}{\ding{51}}
& \textcolor{red}{\ding{55}}
& \textcolor{red}{\ding{55}}
& \textcolor{green!70!black}{\ding{51}} \\
OPA-ext~\cite{liu2021opa}
& 0.15M & 2.2 & 4.3
& \textcolor{green!70!black}{\ding{51}}
& \textcolor{red}{\ding{55}}
& \textcolor{red}{\ding{55}}
& \textcolor{green!70!black}{\ding{51}} \\
PIPE~\cite{wasserman2024paint}
& 0.89M & 1 & 0
& \textcolor{red}{\ding{55}}
& \textcolor{red}{\ding{55}}
& \textcolor{green!70!black}{\ding{51}}
& \textcolor{red}{\ding{55}} \\
SAM-FB~\cite{he2024affordance}
& 3M & 1 & 0
& \textcolor{red}{\ding{55}}
& \textcolor{red}{\ding{55}}
& \textcolor{green!70!black}{\ding{51}}
& \textcolor{red}{\ding{55}} \\
\rowcolor{gray!15}
\textbf{\textsc{HiddenObjects}}
& \textbf{\numannotation}
& \textbf{77.7}
& \textbf{926.3}
& \textcolor{green!70!black}{\ding{51}}
& \textcolor{green!70!black}{\ding{51}}
& \textcolor{green!70!black}{\ding{51}}
& \textcolor{green!70!black}{\ding{51}} \\
\bottomrule
\end{tabular}%
}
\end{table*}

\mypar{Spatial Priors.}
We define a \textit{spatial prior} as the probability distribution of valid bounding box locations for inserting a foreground object into a  scene. Recent approaches learn spatial priors using VLMs that reason about geometric structure~\cite{abdelreheem2025Placeit3d,huang2025fireplace}, while others model domain-specific object priors~\cite{rishubh2025monoplace3D,petersen2025scene}. However, these methods are domain-limited, require 3D annotations, and do not generalize to images. Instead, we learn spatial priors directly from scene images by distilling  implicit object-location knowledge embedded in diffusion models trained on internet-scale image data~\cite{schuhmann2022laion}.

\mypar{Object Placement.}
Object placement identifies suitable locations for inserting a foreground object into a background scene with  semantic and geometric consistency. Early \textit{context-based methods}~\cite{dvornik2018modeling,volokitin2020efficiently} evaluate candidate locations but cannot propose new ones. More recently, \textit{location models} learn placement distributions from manually annotated data~\cite{yun2025imaginingunseengenerativelocation,poska2025hopnet,cheng2025diverse,zhou2022learning,zhang2020and,zhu2023topnet,niu2022fast,zhou2022sac,lee2018context,lin2018st}. However, manual annotation is not scalable, limiting  generalization of models trained on these datasets.  Some datasets generate placements by copy-pasting foreground objects into background scenes without adapting perspective~\cite{liu2021opa,qin2024think}. To overcome these limitations, synthetic approaches~\cite{zhou2025bootplace,winter2025objectmate,gao2025object,wasserman2024paint} generate large-scale training data by removing foreground objects via inpainting. Unlike our approach, these datasets introduce inpainting artifacts (Fig.~\ref{fig:artefact}) into the input background images, enabling shortcut learning.

\mypar{Vision Language Models (VLMs).}
Vision Language Models provide an alternative for object placement. Some methods~\cite{canet2024thinking,he2024affordance,yuan2023learning} generate context descriptions to support background outpainting and create annotated insertion pairs, while others~\cite{li2025freeinsert,liang2025hocomp} directly predict placements. However, VLM-generated context lacks  diversity of large-scale scene datasets~\cite{zhou2017places}, and direct location prediction is often unreliable. More closely related to our work,  methods~\cite{GaoJosephDeLaTorre2025Teleportraits,tewel2025addit,parihar2024text2place,liang2025hocomp,kulal2023putting} exploit implicit spatial priors of diffusion models for placement prediction. However, these priors remain tied to downstream image editing. In contrast, our scalable method distills diffusion-model spatial priors into annotations that model both plausible and implausible placements for downstream applications.

\mypar{Datasets.}
Existing object placement datasets pair an object-free background with a bounding box specifying a placement location (Tab.~\ref{tab:dataset_stats_dense}). They are  constructed in two ways. The first relies on manually annotating placements in real images~\cite{liu2021opa,li2023dreamedit,lu2023dreamcom}. The second removes objects from real scenes using object-removal pipelines to obtain object-free backgrounds~\cite{parihar2024text2place,wasserman2024paint,winter2025objectmate}. Manual annotation is not scalable, while object-removal pipelines introduce artifacts into the input backgrounds. In contrast, our pipeline automatically generates placements for real images.

\section{HiddenObjects Dataset}
\label{sec:method}
The main contribution of this work is a scalable pipeline that automatically extracts \textit{spatial priors} for scene--object pairs on real background images. In Sec.~\ref{framework}, we describe our framework, and in Sec.~\ref{dataprep}, the dataset construction.

\begin{figure*}[t]
\centering
\includegraphics[width=\linewidth]{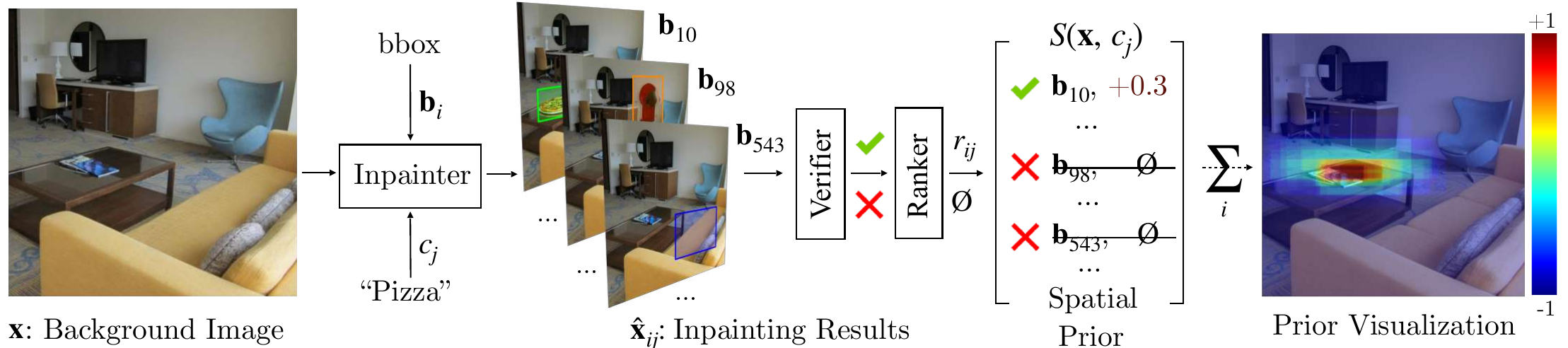}
\caption{
\textbf{Spatial Prior Extraction.}
Given a background image $\mathbf{x}$, the inpainter $\mathcal{I}$ synthesizes an image $\hat{\mathbf{x}}_{ij}$ for target class $c_j$ at bounding box $\mathbf{b}_{i}$. The verifier $\mathcal{V}$ evaluates inpainted images $\hat{\mathbf{x}}_{ij}$, yielding a detected bounding box, and the ranker $\mathcal{R}$ assigns a preference score $r_{ij}$. Repeating this process across sampled boxes yields the ground-truth spatial prior $\mathbf{S}(\mathbf{x}, c_j)$ for the $j$-th class. The heatmap visualizes the aggregated dense distribution of preferences over spatial locations ({\textcolor{black!0!blue}{no}}, {\textcolor{black!20!green}{mid}}, {\textcolor{black!10!red}{good}}).
}
\label{fig:pipeline}
\end{figure*}

\subsection{Spatial Prior Extraction Framework}
\label{framework}
The construction of a spatial prior $\mathbf{S}(\mathbf{x},c)$ between a background image $\mathbf{x}$ and object $c$ consists of three operations (Fig.~\ref{fig:pipeline}). First, an inpainter $\mathcal{I}$ synthesizes class-consistent object insertions in proposed regions. Second, a verifier $\mathcal{V}$ validates that the inpainter correctly added the target object. Third, for each verified insertion, a ranker $\mathcal{R}$ quantifies object-conditioned preference.

\mypar{Spatial Prior.} In the context of our dataset, a spatial prior $\mathbf{S}(\mathbf{x}, c)$ for an input image $\mathbf{x}\in \mathbb{R}^{H\times W\times 3}$ and category $c\in \mathcal{C}$ is defined as a set:
\begin{equation}
\mathbf{S}(\mathbf{x},c) \equiv \{ (\mathbf{b}_{i}, r_{i}) \mid i\in [0, N-1] \}
\end{equation}
where $r_{i} \in \mathbb{R}$ is the score of a corresponding bounding box $\mathbf{b}_{i}=(x_i,y_i,w_i,h_i)$, with box center $(x_i,y_i)$ and size $(w_i,h_i)$ for a fixed set of $N$ boxes. We define a deterministic set of boxes $\mathbf{b}_i$ produced by a sliding window at different resolutions that evenly span the whole image (more details in 
Appendix~\ref{supmat:bbox}. For simplicity, all images have the same resolution.

\mypar{Inpainter.} The goal of the inpainter $\mathcal{I}$ is to produce insertion proposals at different locations in the input image. We denote the inpainting result:
\begin{equation}
\mathbf{\hat{x}}_{ij}\equiv \mathcal{I}(\mathbf{x}, \mathbf{b}_i, c_j),
\end{equation}
in the region defined by bounding box $\mathbf{b}_i$ for class $c_j$.  It is possible that $\exists\, i,j: \hat{\mathbf{x}}_{ij} \approx \mathbf{x}$ or that the inpainting does not contain $c_j$ (see Fig.~\ref{fig:teaser} \textbf{(a)}). Such failures are filtered out by the verifier.

\mypar{Verifier.} The goal of the verifier $\mathcal{V}$ is to identify whether the target category has been correctly inpainted. Concretely, given a synthesized image $\hat{\mathbf{x}}_{ij}$ for input bounding box $\mathbf{b}_i$ and target class $c_j$, we use a conditional verifier $\mathcal{V}(\mathbf{x}, c_j)$ that produces a set of detections $\mathbf{B}_{ij}$:
\begin{equation}
\mathbf{B}_{ij} \equiv \{\mathbf{b} \mid (\mathbf{b}, \psi) \in \mathcal{V}(\hat{\mathbf{x}}_{ij}, c_j),\; \psi \geq \tau \}
\end{equation}
\noindent where each $\mathbf{b}$ is predicted with confidence $\psi \in [0, 1]$ and $\tau \in [0,1]$ is the confidence threshold. When $|\mathbf{B}_{ij}| = 0$, we consider the prediction as negative by setting $\mathbf{b}_{ij} = \varnothing$. Otherwise, we select the verifier bounding box $\mathbf{b}$ with the highest \textbf{I}ntersection \textbf{o}ver \textbf{U}nion (IoU)  with the input bounding box $\mathbf{b}_i$:
\begin{equation}
\mathbf{b}_{ij} = \arg\max_{\mathbf{b} \in \mathbf{B}_{ij}} \mathrm{IoU}(\mathbf{b}, \mathbf{b}_i)
\end{equation}
\noindent This matches the detection to the inpainted region $\mathbf{b}_i$. Instead of adding objects, inpainting may replace objects already present in the background image $\mathbf{x}$. To guarantee that no placement corresponds to a pre-existing object of $c_j$, we set $\mathbf{b}_{ij} = \varnothing$ if the verified bounding box $\mathbf{b}_{ij}$ has IoU $\geq 0.5$ with verified objects of the same class in $\mathbf{x}$.

\begin{figure}[t]
\centering
\includegraphics[width=\linewidth]{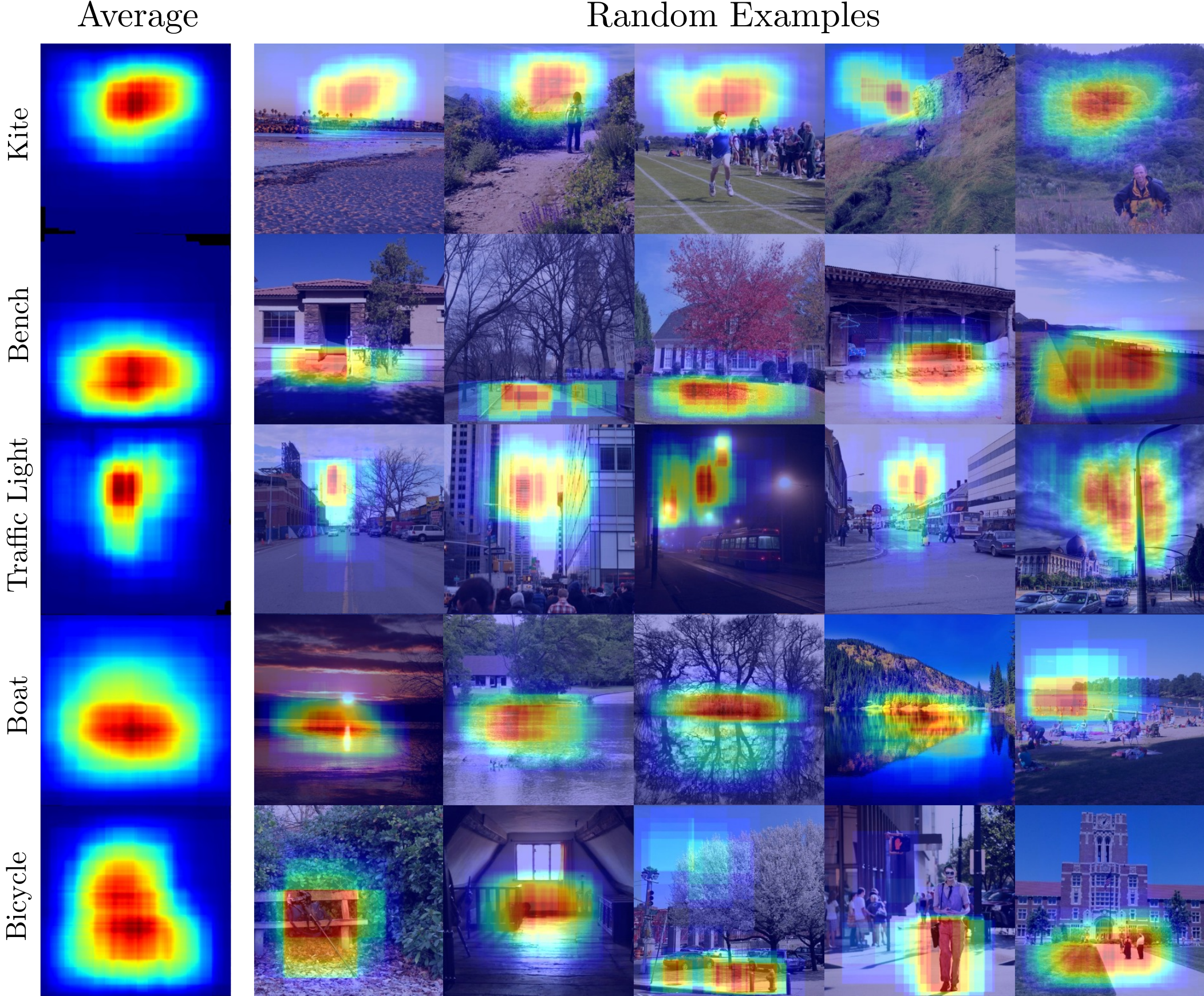}
\caption{\textbf{Spatial Prior (\textsc{HiddenObjects})} aggregated as a heatmap. (\textbf{Left}) Mean spatial priors across all \textsc{HiddenObjects} images for the object classes shown. Classes such as \textit{bench} and \textit{boat} appear predominantly near the bottom, while \textit{kite} is more frequent near the top. These patterns reflect semantic regularities (e.g., kites in the sky) and photographic biases in image datasets. (\textbf{Right}) Examples of per-instance priors. Extended in Appendix~\ref{sec:supp_prior_viz}.}
\label{fig:cherry_picked_dataset}
\end{figure}

\mypar{Ranker.} After verifying insertions, a conditional ranker $\mathcal{R}(\mathbf{x}, c)$ ranks which insertions are preferred. This is essential for constructing the final object placement prior, as it quantifies preference across valid placement locations.  For each image $\hat{\mathbf{x}}_{ij}$ and class $c_j$, the final extracted prior $\mathbf{S}(\mathbf{x},c_j)$ is:
\begin{equation}
\begin{aligned}
\mathbf{S}(\mathbf{x},c_j) &= \{(\mathbf{b}_{ij}, r_{ij}) \mid \mathbf{b}_{ij} \neq \varnothing \}, \\
&\quad \text{where } r_{ij} \equiv \mathcal{R}(\hat{\mathbf{x}}_{ij}, c_j).
\end{aligned}
\label{eq:reward}
\end{equation}

\subsection{Dataset Construction}
\label{dataprep}

To construct \textsc{HiddenObjects}, we select images from Places365~\cite{zhou2017places} for its diversity of scenes. We randomly select 126 unique background classes. Then we pick 10 macro object categories and, for each, select 5 classes from COCO~\cite{lin2014microsoft}, resulting in 50 distinct object categories $c_j$. To avoid implausible insertions, we manually pair object categories with background categories (see Appendix~\ref{app:bgfg}).
Given a set of valid background and foreground object pairs, we randomly sample combinations until we compute 27K spatial priors, with $1,004$ bounding boxes each. After the ranker stage, the final dataset consists of \numscenes positively annotated backgrounds and 2,097,587 positive bounding box annotations. \textsc{HiddenObjects} is split into train (85\%), validation (10\%), and test (5\%) sets.

\mypar{Framework Implementation.} Unless stated otherwise, our pipeline uses Qwen-Image~\cite{wu2025qwen} with a ControlNet adaptor\footnote{\url{https://huggingface.co/InstantX/Qwen-Image-ControlNet-Inpainting}} (Qwen+CN) as the inpainting backbone, with 20 inference steps and a true classifier-free guidance scale of 4.0. As verifier, we use Grounded-SAM-2~\cite{ren2024grounded} for open-set object detection with confidence threshold $\tau=0.4$, and ImageReward~\cite{xu2023imagereward} as the preference ranker. Ablations are in Sec.~\ref{sec:key_componenet_selection}.

\mypar{Speedup.}
\label{sec:speedup_text} For most insertion proposals, we expect the inpainting model to produce missing or wrong objects. Given that the diffusion inpainter is slow, detecting faulty inpaintings early during denoising would avoid unnecessary computational costs.
We achieve this by measuring how much the noise predicted by the denoising network $\varepsilon_{\theta}$ varies between the conditional $\varepsilon_{\theta}(x_t, c_j)$ and unconditional $\varepsilon_{\theta}(x_t, \varnothing)$ branches at each denoising step $x_t$.  We find that the L1-difference between the two branches correlates highly with prompt adherence, with successful generations exhibiting higher values than failed generations.  We compute a threshold that detects faulty boxes with 80.4\% recall after two denoising steps, leaving only 34\% of proposals for inpainting and yielding a 2.2$\times$ speedup without degrading prior quality. More detailed analysis in Appendix~\ref{sec:supp_speedup}.

\section{Pipeline Validation and Dataset Analysis}
\label{sec:experiments}

We evaluate our diffusion-based spatial prior extraction pipeline along several axes:
(\S\ref{sec:key_componenet_selection}) key component selection,
(\S\ref{sec:data_quality}) validation of pipeline labels,
(\S\ref{sec:downstrea_image_editing_performance}) downstream image editing quality,
and (\S\ref{sec:systematic_biases}) systematic bias analysis.

\subsection{Key Component Selection} 
\label{sec:key_componenet_selection} 
Here, we justify the selection of the two main components: the inpainter $\mathcal{I}(\mathbf{x}, \mathbf{b}_i, c_j)$ and the ranker $\mathcal{R}(\mathbf{\hat{x}}_{ij}, c_j)$.

\mypar{Experimental Setup.}
We evaluate component choices on OPA~\cite{liu2021opa}, the largest
placement dataset with human annotations for both positive and negative
placements. We restrict it to classes shared with \textsc{HiddenObjects},
yielding 464 backgrounds, and run our pipeline to produce
1{,}004 placement annotations per image. To compare against human judgments,
we match each OPA annotation (positive or negative) to its closest pipeline annotation by IoU.

\begin{table}[!t]
    \centering
    \small
    \setlength{\tabcolsep}{2pt}
    \caption{\textbf{Component Analysis.} \textbf{(Left)} We evaluate different rankers for the same inpainter (FLUX). \textbf{(Right)} We evaluate different inpainters for the same ranker (ImageReward).
    ImageReward and Qwen perform best.  CN: ControlNet.}
    \label{tab:component_selection}
    \resizebox{\linewidth}{!}{
        \begin{minipage}[t]{0.48\linewidth}
            \centering
            \begin{tabular}{lc}
                \toprule
                 \textbf{Ranker} & \textbf{AP} $\uparrow$ \\
                \midrule
                 Random & 52.8 \\
                 Aesthetics & 52.6 \\
                 CLIP Score & 53.8 \\
               \rowcolor{gray!15}    \textbf{ImageReward} & \textbf{57.1} \\
                \bottomrule
            \end{tabular}
        \end{minipage}
        \hfill
        \begin{minipage}[t]{0.51\linewidth}
            \centering
            \begin{tabular}{ll}
                \toprule
                \textbf{Inpainter}  & \textbf{F1} $\uparrow$ \\
                \midrule
                SDXL &  52.6 \\
                SDXL + CN &  49.8 \\
                FLUX &  48.5 \\
                FLUX + CN & 40.1 \\
              \rowcolor{gray!15}    \textbf{Qwen + CN} &  \textbf{53.5} \\
                \bottomrule
            \end{tabular}
        \end{minipage}
    }
    \vspace{-1em}
\end{table}

\mypar{Ranker Selection.} 
To score verifier-accepted placements, we compare ImageReward~\cite{xu2023imagereward}, Aesthetic Score~\cite{aestheticpredictor2024}, and CLIP Score~\cite{hessel2021clipscore}. %
For a fair comparison, we fix the inpainter to FLUX~\cite{batifol2025flux} across all methods. In Tab.~\ref{tab:component_selection}, we evaluate each ranker’s ability to prioritize valid over invalid insertions using Average Precision (AP). ImageReward achieves the highest AP (57.1\%), outperforming CLIP and Aesthetics. We select  ImageReward as our ranker.

\mypar{Inpainter Selection.} 
We evaluate different inpainter backbones: U-Net based diffusion (SDXL~\cite{podell2023sdxl}) and DiT flow-matching (FLUX~\cite{batifol2025flux}, Qwen+CN).  
We see in Tab.~\ref{tab:component_selection} that Qwen+CN offers best alignment to sparse human labels via highest F1-score (53.5), which we select for future experiments.
See  Appendix~\ref{sec:supp_component_qualitative} for extended  comparisons.

\subsection{Human Validation of Placement Labels}
\label{sec:data_quality}

We audit whether placement labels produced by our pipeline agree with human
judgments of plausibility.

\noindent\textbf{Experimental Setup.}
We randomly sample 50 \textsc{HiddenObjects} test scenes, each with 1,004 verifier-labeled placements. Annotators view a $3\times3$ grid containing the three verifier-accepted placements with the highest ImageReward scores and six uniformly sampled verifier-rejected placements, overlaid on the original background, and select the three most plausible boxes for the target class. The study comprises 505 responses from 10 annotators, covering 450 distinct boxes (150 passed, 300 rejected) and 4,545 box-level judgments. See Appendix~\ref{sec:supp_human_eval} for examples.

\noindent\textbf{Evaluation Protocol.}
For each box, human acceptance is the fraction of annotators selecting it. At the grid level, we report pipeline--human overlap (fraction of selected boxes that are verifier-accepted) and mean pairwise human--human overlap. At the box level, we compare acceptance scores for passed vs.\ rejected boxes and report ROC--AUC. Thresholding acceptance at 50\% (ties positive), we evaluate verifier predictions using accuracy, precision, recall, specificity, $F_1$, and Cohen's $\kappa$, and Fleiss' $\kappa$ for inter-annotator agreement.

\begin{figure}[t]
\centering
\includegraphics[width=\columnwidth]
{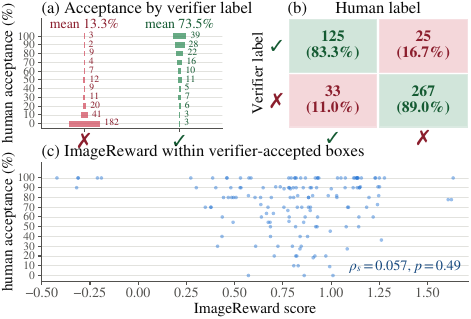}
\caption{\textbf{Analysis of placement labels.}
(a) Human acceptance is substantially higher for verifier-accepted than
verifier-rejected boxes (ROC--AUC is $0.931$).
(b) Human labels have 87\% agreement with verifier's labels.
(c) ImageReward has negligible human acceptance correlation on verifier-accepted placements.}
\label{fig:studyB_box_analysis}
\end{figure}

  \noindent\textbf{Human Alignment Results.}
  At the \textit{grid level}, verifier-accepted placements account for $73.3\%$
  of annotator selections, comparable to the $72.5\%$ mean pairwise overlap
  between annotators. At the \textit{box level}, verifier-accepted placements
  receive substantially higher mean human acceptance than verifier-rejected
  placements ($0.735$ vs.\ $0.133$; Fig.~\ref{fig:studyB_box_analysis}a),
  yielding an ROC--AUC of $0.931$. Against the thresholded human reference,
  the verifier achieves $87.1\%$ accuracy, $83.3\%$ precision, $79.1\%$
  recall, $91.4\%$ specificity, and an $F_1$ score of $81.2\%$, with
  substantial agreement beyond chance (Cohen's $\kappa=0.714$;
  Fig.~\ref{fig:studyB_box_analysis}b). Human judgments show moderate
  inter-annotator agreement (Fleiss' $\kappa=0.593$), with unanimous judgments
  for $50.4\%$ of boxes. The 58 verifier--human disagreements among 450 boxes
  (systematic error rate of $12.9\%$) comprise 25 false positives (verifier-accepted but
  human-implausible) and 33 false negatives (verifier-rejected but
  human-plausible). Overall, these results demonstrate strong pipeline--human
  alignment.
  
\noindent\textbf{Ranker Analysis.}
ImageReward ranks verifier-accepted candidates for top-3 selection (98.6th percentile). 
Even among the top-ranked subset, ImageReward has negligible correlation with human acceptance ($\rho_s=0.057$, $p=0.49$; Fig.~\ref{fig:studyB_box_analysis}c). We therefore use ImageReward as a top-$k$ selection heuristic, not a calibrated plausibility score.

\subsection{Downstream Image Editing Performance}
\label{sec:downstrea_image_editing_performance}
Having established the optimal pipeline components and validated our labels, we now evaluate whether the extracted priors improve downstream object insertion
 performance. 

\begin{figure*}[t]
    \centering
    \includegraphics[width=\linewidth]{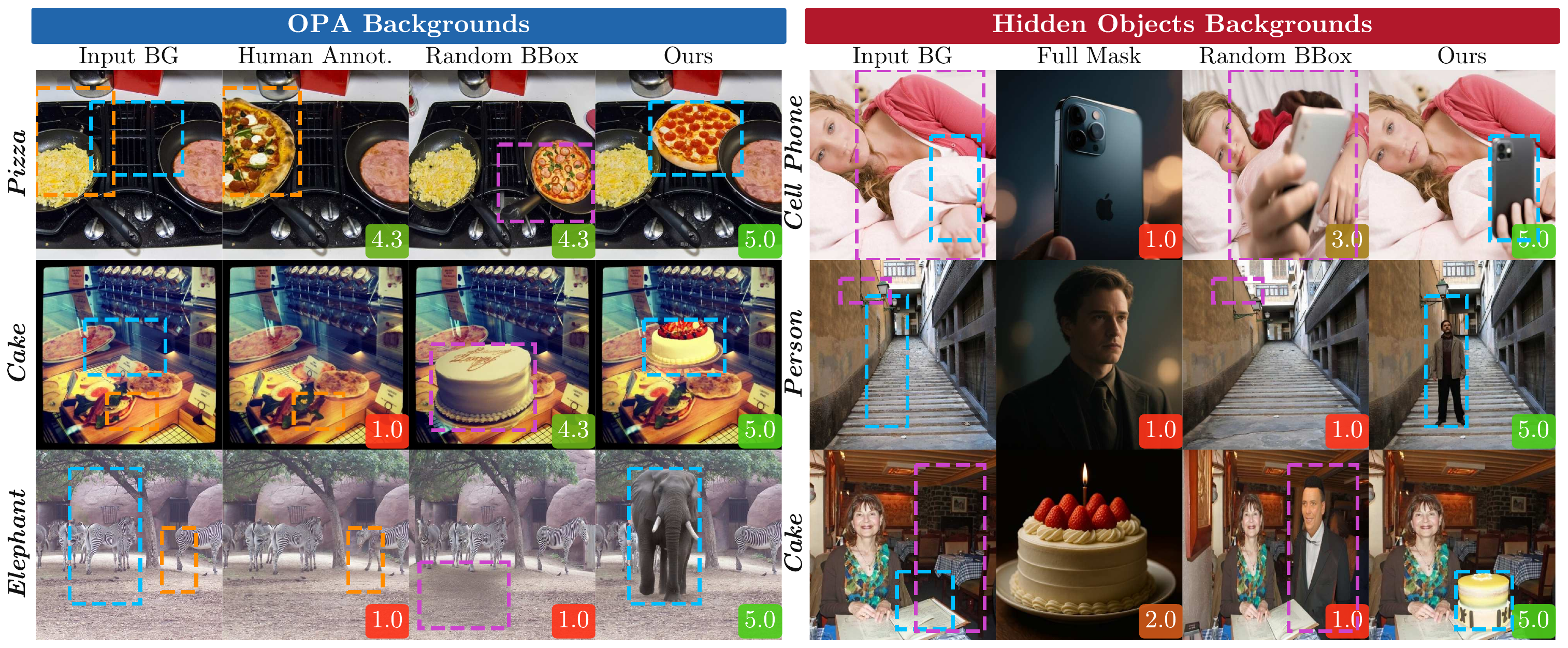}
\caption{
\textbf{Image inpainting with object placement priors.}
We evaluate four object placement approaches on the \textsc{OPA}~\cite{liu2021opa} and
\textsc{HiddenObjects} backgrounds.
\emph{Full Mask} inpaints the entire image.
\emph{Human Annot.}\ uses ground-truths from OPA.
\emph{Random BBox} samples a uniformly random placement region, serving as a
controlled baseline.
\emph{Ours} uses the top-ranked box of our dataset.
Dashed boxes show the inpainting region
(\textcolor[HTML]{00BFFF}{HiddenObject annotation (ours)},
 \textcolor[HTML]{FF8C00}{OPA annotation},
 \textcolor[HTML]{CC44CC}{Random}).
Scores are (1--5) ImgEdit-Judge~\cite{ye2025imgedit}.
Our pipeline consistently produces well-placed, photorealistic composites
across object classes and image scenes, outperforming sparse annotations.
}
    \label{fig:qualitative_imgedit}
\end{figure*}

 \begin{table}[t]
  \centering
  \small
  \caption{\textbf{Downstream image-editing quality} measured by
  ImgEdit-Judge~\cite{ye2025imgedit} (1--5 $\uparrow$; higher is better).}
  \label{tab:downstream_imgedit}
  \setlength{\tabcolsep}{5pt}
  \begin{tabular}{@{}l l@{\hspace{2pt}}l l@{\hspace{2pt}}l@{}}
  \toprule
  \textbf{Source}
  & \multicolumn{2}{c}{\textbf{\textsc{HiddenObjects}}}
  & \multicolumn{2}{c}{\textbf{\textsc{OPA}}} \\
  \midrule
  Raw Background
  & 1.04 & & 1.00 & \\
  Full Mask
  & 1.60 & & 1.66 & \\
  BootPlace~\cite{zhou2025bootplace}
  & 1.17 & & 1.01 & \\
  OPA Human Annotation
  & --   & & 2.68 & \\
  PlaceNet~\cite{zhang2020learning}
  & 2.63 & & 2.94 & \\
  Random Box
  & 2.65 & & 2.89 & \\
  TerseNet~\cite{tripathi2019learning}
  & 2.99 & & 3.03 & \\
  GracoNet~\cite{zhou2022learning}
  & 3.19 & & 3.18 & \\
  Ours (Aesthetic)
  & 3.35 & {\scriptsize\color{red}$-0.34$}
  & 3.51 & {\scriptsize\color{red}$-0.39$} \\
  Ours (CLIP Score)
  & 3.61 & {\scriptsize\color{red}$-0.08$}
  & 3.84 & {\scriptsize\color{red}$-0.06$} \\
  \rowcolor{gray!20}
  \textbf{\rule[-0.6ex]{0pt}{2.6ex}Ours (ImageReward)}
  & \multicolumn{2}{l}{\textbf{3.69}}
  & \multicolumn{2}{l@{}}{\textbf{3.90}} \\
  \bottomrule
  \end{tabular}
  \end{table}
\mypar{Evaluation Setup}. We evaluate object placement in both \textsc{HiddenObjects} and OPA test sets. For fair comparison, we take a subset of backgrounds that have the same 28 shared COCO classes. After filtering, we select 200 random images from \textsc{HiddenObjects} and all 76 OPA images. To test image-editing performance, we use the benchmark ImgEdit-Judge~\cite{ye2025imgedit}, which relies on Qwen2.5-VL~\cite{bai2025qwen25vltechnicalreport} to evaluate image quality by assigning a score between 1 and 5. This evaluation measures bounding box suitability for diffusion-based insertion, not ground-truth plausibility. 

\mypar{Quantitative Analysis.} In Tab.~\ref{tab:downstream_imgedit}, we compare inpainting performance on both test sets against four types of baselines: (i) no edit, where the raw background is passed directly; (ii) full mask, covering the entire image; (iii) random bounding box, uniformly sampled with area between 10\% and 90\% of the image and aspect ratio between 0.25 and 4; and (iv) placement models (BootPlace, PlaceNet, TerseNet, GracoNet). In OPA, we also compare with inpainting from sparse annotations. Our method outperforms all baselines, indicating that our annotations identify boxes better aligned for context-aware diffusion-based synthesis.
Random boxes score higher than sparse labels, suggesting ImgEdit-Judge partly rewards box size and motivating further human evaluation.

\mypar{Human Evaluation.}
We extend our analysis by further verifying the findings of Tab.~\ref{tab:downstream_imgedit} (3.90 ours annotation vs.\ 2.68 OPA annotation) through direct human evaluation rather than VLM-as-a-judge metrics. We show human evaluators randomized side-by-side inpainting pairs generated using placements from either sparse OPA annotations or our priors, on the same 76 OPA-test backgrounds as Tab.~\ref{tab:downstream_imgedit}. See Appendix~\ref{sec:supp_image_editing} for examples. Across 1055 responses from 15 annotators, inpaintings generated using our priors achieved a \textbf{65.6\%} human preference rate over those generated using sparse human OPA annotations. This result provides independent human validation that the improvements observed in Tab.~\ref{tab:downstream_imgedit} correspond to visible differences in generation quality and our extracted priors provide consistent guidance signals for downstream inpainting.

\mypar{Ranker Sensitivity Analysis.}
Replacing ImageReward with CLIP or Aesthetic Score changes the selected box (top-1) in 81--95\% of samples (mean IoU 0.35--0.44), yet ImgEdit-Judge drops by only 0.08/0.06 and 0.34/0.39 on \textsc{HiddenObjects}/OPA, respectively (Tab.~\ref{tab:downstream_imgedit}). All variants still outperform sparse OPA annotations. Finally, the ranking ImageReward $>$ CLIP $>$ Aesthetic matches Tab.~\ref{tab:component_selection}, further supporting our ranker choice (Sec.~\ref{sec:key_componenet_selection}) and indicating that placement quality is driven primarily by the verifier (Sec.~\ref{sec:data_quality}).

\subsection{Systematic Biases Analysis}
\label{sec:systematic_biases}
\begin{figure*}[t]
\centering
\begin{subfigure}[b]{0.47\textwidth}
\centering
\includegraphics[height=3.5cm]{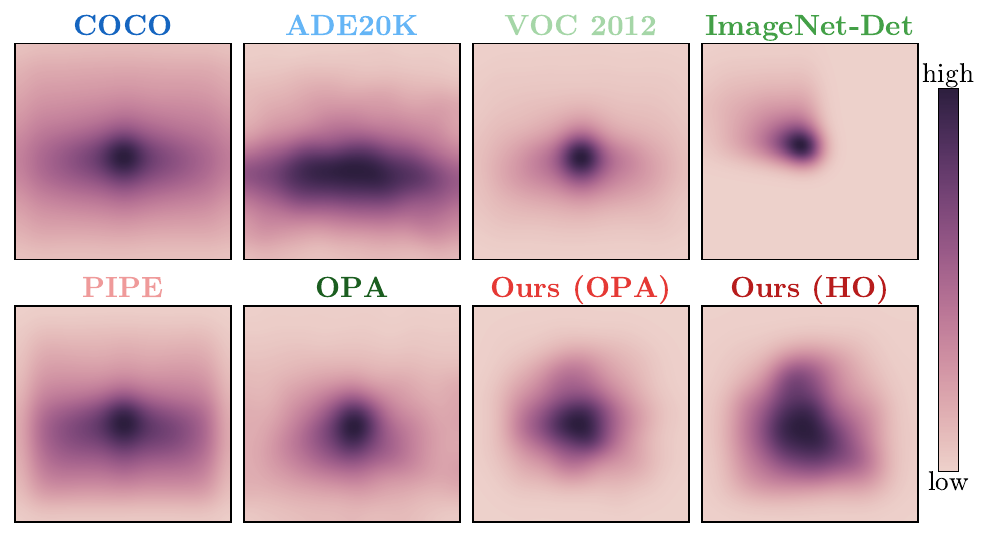}
\caption{Distribution of Object Centers}
\end{subfigure}
\hfill
\begin{subfigure}[b]{0.47\textwidth}
\centering
\includegraphics[height=3.5cm]{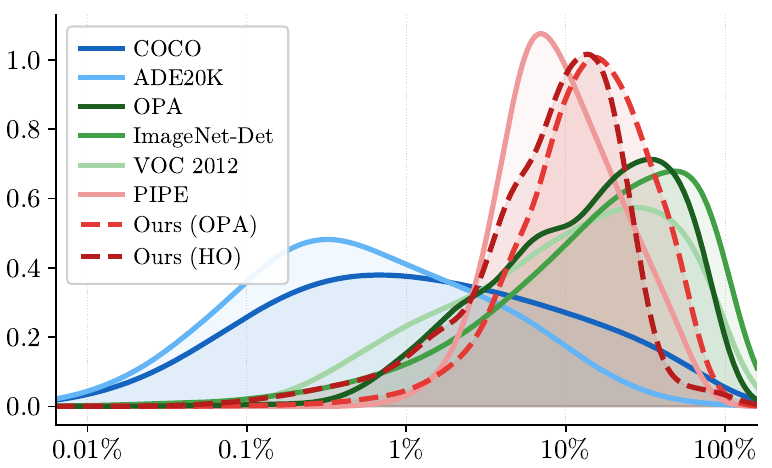}
\caption{Density of Relative Object Area}
\end{subfigure}
\caption{\textbf{Analysis of Spatial and Scale Distributions.}
\textbf{(a)} Object center distributions across datasets: while benchmarks like COCO~\cite{lin2014microsoft}, VOC~\cite{pascal-voc-2012}, and ImageNet~\cite{imagenet} exhibit a pronounced center-crop bias, our pipeline (evaluated on \textsc{HiddenObjects} and OPA backgrounds) yields a more diffuse distribution with higher spatial variance.
\textbf{(b)} Distribution of bounding box areas: \textsc{HiddenObjects} and PIPE~\cite{wasserman2024paint} lack small objects (0.01\%--0.1\% image area), revealing a shared bottleneck of underlying diffusion backbones in inpainting small areas.}
\label{fig:distribution_of_boxes}
\end{figure*}

A key challenge in object placement is that training datasets encode spatial and scale biases that models inevitably learn. While our extracted spatial priors capture human-aligned placement expectations, they also inherit systematic biases from the diffusion model and verifier training data. We therefore compare our distribution with object detection datasets, including COCO~\cite{lin2014microsoft}, VOC 2012~\cite{pascal-voc-2012}, ADE20K~\cite{zhou2019semantic}, ImageNet-Det~\cite{imagenet}, and placement datasets such as PIPE~\cite{wasserman2024paint} and OPA~\cite{liu2021opa}.

\mypar{Distribution of Object Centers.}
Fig.~\ref{fig:distribution_of_boxes}a shows the distribution of bounding box centers (similar to~\cite{gupta2019lvis}).  COCO~\cite{lin2014microsoft}, VOC 2012~\cite{pascal-voc-2012}, PIPE~\cite{wasserman2024paint}, and ImageNet-Det~\cite{imagenet} exhibit a pronounced center bias, concentrating objects near the image center. This reflects photographers' tendency to center subjects. In contrast, our dataset shows the most spatially uniform distribution of object placement among object-centric recognition datasets. Still, unlike ADE20K, our spatial priors retain the compositional biases of the large-scale web image datasets used to train diffusion models. For example,  in Fig.~\ref{fig:cherry_picked_dataset}, although a bench could be placed anywhere on the ground, the prior exhibits a bottom-center bias.

\mypar{Distribution of Object Scale.}
The object size distribution (Fig.~\ref{fig:distribution_of_boxes}b) reveals another difference. Object-centric datasets (VOC~\cite{pascal-voc-2012} and ImageNet-Det~\cite{imagenet}) are dominated by large objects occupying 10--100\% of the image area. In contrast, our pipeline favors smaller sizes (1--10\%), which reflect realistic object placement in complex scenes. Despite this shift, the extracted priors remain biased toward medium-to-large bounding boxes. This results from a limitation of diffusion models, which struggle to produce quality inpaintings in very small regions. This resolution limit introduces a perspective bias. In high-depth scenes, objects near the vanishing point occupy fewer pixels, causing inpaintings to fail the verifier's detection threshold.

\section{Model Distillation for Object Placement}
\label{sec:model_distillation}
\label{sec:benchmarking_efficiency}

\begin{table*}[t]
  \centering
  \small
  \caption{Evaluation of placement methods on the
  \textsc{HiddenObjects} test set (\%). mAP$_\text{img}$ is AP at IoU $0.50$ computed per sample and averaged over all 545 samples; mAP$_\text{cls}$ is PASCAL VOC 11-point AP computed per class and averaged over the 28 classes.}
  \label{tab:baseline_comparison}
  \resizebox{\textwidth}{!}{%
  \begin{tabular}{llrrrrrr}
    \toprule
    \textbf{Method} & \textbf{Train Set} & \textbf{mAP$_\text{img}$} $\uparrow$ & \textbf{mAP$_\text{cls}$} $\uparrow$ & \textbf{IoU50@1} $\uparrow$ & \textbf{IoU@1} $\uparrow$ & \textbf{IoU50@5} $\uparrow$ & \textbf{IoU@5} $\uparrow$ \\
    \midrule
    \modelicon{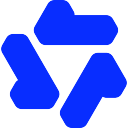}Qwen2.5-VL-72B~\cite{bai2025qwen25vltechnicalreport} & \textsc{Zero-shot} & 1.3 & 2.6 & 7.9 & 14.0 & 12.7 & 18.8 \\
    \modelicon{qwen.png}Qwen2.5-VL-7B & \textsc{Zero-shot} & 1.4 & 2.2 & 8.6 & 13.9 & 10.1 & 16.5 \\
    \modelicon{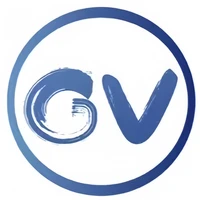}InternVL3-8B~\cite{chen2024internvl} & \textsc{Zero-shot} & 0.3 & 0.7 & 3.3 & 11.2 & 9.2 & 18.7 \\
    \modelicon{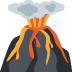}LLaVA-OneVision-7B & \textsc{Zero-shot} & 1.0 & 3.0 & 11.7 & 20.9 & 16.3 & 24.7 \\
    \midrule
    BootPlace~\cite{zhou2025bootplace} & \textsc{Cityscapes} & 0.1 & 0.1 & 0.9 & 6.0 & 0.9 & 6.0 \\
    TerseNet~\cite{tripathi2019learning} & \textsc{OPA} & 1.4 & 4.0 & 20.9 & 29.5 & 20.9 & 29.5 \\
    GracoNet~\cite{zhou2022learning} & \textsc{OPA} & 2.1 & 3.9 & 23.7 & 35.9 & 23.7 & 35.9 \\
    PlaceNet~\cite{zhang2020learning} & \textsc{OPA} & 2.9 & 5.4 & 43.5 & 42.4 & 43.5 & 42.4 \\
    \midrule
    Ours & \textsc{OPA} & 28.0 & 7.8 & 33.0 & 36.4 & 57.6 & 51.9 \\
    \rowcolor{gray!15}
    \textbf{Ours} & \textbf{\textsc{HiddenObjects}} & \textbf{56.6} & \textbf{21.0} & \textbf{62.9} & \textbf{55.2} & \textbf{79.1} & \textbf{67.7} \\
    \bottomrule
  \end{tabular}%
  }
\end{table*}

To infer placements efficiently, we distill our prior into a lightweight transformer conditioned on the background and object class. See Appendix~\ref{sec:supp_detr_impl} for implementation details.

\mypar{Experimental Setup.} We evaluate our model on OPA and \textsc{HiddenObjects}. For fair cross-dataset evaluation, we use the subset of \textsc{HiddenObjects} containing the 28 foreground object classes shared with existing placement datasets, resulting in 9,438 training and 545 test samples (565,948 location annotations). Similarly, we train on OPA 28 shared classes (958 training images). We evaluate bounding box proposals using Intersection over Union (IoU) and mean average precision (mAP) at an IoU threshold of $0.50$, and report IoU50 accuracy, where predictions with a maximum IoU $\geq 0.50$ to the ground truth are counted as positive hits. 

\mypar{Quantitative Evaluation.} We benchmark our distilled model against state-of-the-art generalist VLMs and dedicated placement networks. In Tab.~\ref{tab:baseline_comparison}, we observe that the strongest zero-shot VLM peaks at 3.0\% mAP$_\text{cls}$ (for prompting details see Appendix~\ref{sec:supp_vlm_placement_protocol}). Existing placement models struggle beyond top-1 localization because they are trained only on sparse spatial priors. 
In contrast, our model performs best with 21.0\% mAP$_\text{cls}$, 62.9\% IoU50@1, and 79.1\% IoU50@5.   Similar to Sec.~\ref{sec:downstrea_image_editing_performance}, we evaluate the model's performance in downstream image editing (Tab.~\ref{tab:results_opa_short}).
We show our distilled model produces high-quality placement proposals, outperforming sparse OPA annotations (3.43 vs.\ 2.68).

\mypar{Human Evaluation.}
We conduct A/B preference studies with six evaluators, who compare two placement annotations, without inpainted objects, on clean backgrounds (see Appendix~\ref{sec:supp_human_eval}). We
  collect 474 votes across all 79 eligible OPA scenes. By majority vote per scene, our distilled model is preferred in 38.0\% of cases, the sparse human annotation in 35.4\%, and 26.6\% are ties. Against
  PlaceNet, the most competitive baseline, we collect 120 votes across 20
  randomly selected \textsc{HiddenObjects} scenes. By majority vote, our model is
  preferred on 35.0\%, PlaceNet on 15.0\%, and 50.0\% are ties.
  Overall, our distilled model matches human annotation preference, while outperforming the strongest trained baseline.

\begin{table}[t]
\centering
\caption{Downstream image editing quality on OPA benchmark.}
\label{tab:results_opa_short}
\small %
\setlength{\tabcolsep}{4pt} %
\begin{tabular}{@{}llc@{}}
\toprule
& \textbf{Method} & \textbf{ImgEdit-Judge} $\uparrow$  \\
\midrule
& OPA Annotation & 2.68 \\
\midrule
\multirow{3}{*}{\rotatebox[origin=c]{90}{\small\textbf{Ours}}}
& Model trained on OPA & 3.35 \\
& Model trained on \textsc{HiddenObjects} & 3.43 \\
& \textbf{Diffusion Annotation (Sec.~\ref{sec:method})} & \textbf{3.90} \\
\bottomrule
\end{tabular}
\end{table}

\mypar{Inference Time.}
On an H100 GPU, the distilled model runs at 3.77\, ms/image with 188.9\, MB peak GPU memory. Compared to the full pipeline (including the early-exit speedup in Sec.~\ref{dataprep}) (14.49\, min/image, 65.9\, GB), this is a 230,000$\times$ speedup and a 300$\times$ reduction in memory.

\subsection{Generalization}
\label{sec:5.2}

We evaluate generalization of the distilled model across unseen object categories and background domains.

\mypar{Open-Vocabulary Performance.}
We replace the learned class embeddings with frozen CLIP (ViT-B/32) text embeddings. Trained on the 28 base classes, the model is evaluated zero-shot on 5 unseen classes. 
In Tab.~\ref{tab:openset_comparison}, it achieves 42.4\% mAP$_\text{img}$ on novel categories, retaining 80.3\% of the fully supervised upper bound (52.8\%  mAP$_\text{img}$), indicating good generalization abilities to unseen categories.

\begin{table}[t]
\vspace{-0.6em}
\centering
\caption{\textbf{Open-vocabulary placement on \textsc{HiddenObjects}.} The distilled model is trained on the 28 base classes. Novel classes are evaluated zero-shot using CLIP text embeddings.}
\label{tab:openset_comparison}
\resizebox{\linewidth}{!}{
\setlength{\tabcolsep}{4pt}
\begin{tabular}{lllrrr}
\toprule
\textbf{Setting} & \textbf{Train} & \textbf{Test} & \textbf{mAP$_\text{img}$} & \textbf{IoU50@1} & \textbf{IoU@1} \\
\midrule
Closed-set & Base 28 & Base 28 & 56.6 & 62.9 & 55.2 \\
\midrule
\multirow{2}{*}{Open-vocabulary}
& Base 28 & Base 28 & 57.2 & 66.2 & 57.6 \\
& & Novel 5 & \textbf{42.4} & 52.9 & 48.8 \\
\cmidrule{2-6}
\multirow{2}{*}{Fully supervised}
& Base 28 + Novel 5 & Base 28 & 57.3 & 65.9 & 56.9 \\
& & Novel 5 & \textbf{52.8} & 62.2 & 54.1 \\
\bottomrule
\end{tabular}
}
\vspace{-0.8em}
\end{table}

\mypar{Cross-Dataset Generalization.}
We further evaluate the distilled model on the independent MureCom benchmark~\cite{lu2023dreamcom} . Replacing the original bounding boxes with our predictions improves ImgEdit-Judge from 2.82 to 3.65, demonstrating transfer across background datasets.

\section{Conclusion}
We presented a scalable framework for extracting class-conditioned spatial priors from diffusion models, overcoming sparse manual annotations and artifact-prone synthetic datasets. Our pipeline produces \textsc{HiddenObjects}, a dataset of 27M evaluated placements across 27K backgrounds with human-aligned placement statistics. Distilling these priors into a lightweight model enables real-time inference while cutting computation by five orders of magnitude. 
We outperform both dedicated placement models and zero-shot VLMs, and surpass sparse annotations in downstream image editing quality.
Although the priors inherit photographic composition biases from their generative supervisors, they enforce complex structural and scene constraints. We believe they can advance image editing and scene understanding.

\section*{Acknowledgments}
D. Papadopoulos was supported by
the DFF Sapere Aude Starting Grant “ACHILLES”. S. Belongie and I. Siglidis are supported by the Pioneer Centre for AI, DNRF grant number P1.
We would like to thank Mehmet Onurcan Kaya and Jens Parslov for proofreading and insightful discussions.
{
    \small
    \bibliographystyle{ieeenat_fullname}
    \bibliography{main}
}
\clearpage

\newpage

\setcounter{page}{1}
\renewcommand{\thepage}{A-\arabic{page}}
\definecolor{clr_person}{HTML}{E8F0FE}  
\definecolor{clr_vehicle}{HTML}{F1F8E9} 
\definecolor{clr_outdoor}{HTML}{FFFDE7} 
\definecolor{clr_animal}{HTML}{F3E5F5}  
\definecolor{clr_sports}{HTML}{E0F2F1}  
\definecolor{clr_kitchen}{HTML}{FFF3E0} 
\definecolor{clr_food}{HTML}{FCE4EC}    
\definecolor{clr_furni}{HTML}{EFEBE9}   
\definecolor{clr_tech}{HTML}{E8EAF6}    
\definecolor{clr_misc}{HTML}{ECEFF1}    
\definecolor{myblue}{HTML}{3B6BA5}
\definecolor{figegreen}{RGB}{34, 139, 34} 
\definecolor{figered}{RGB}{220, 20, 60}
\setcounter{section}{0}
\renewcommand{\thesection}{\Alph{section}}

\twocolumn[
\begin{center}
{\LARGE \textbf{Appendix}}
\end{center}
\vspace{1em}
]

\begin{figure}[h]
    \centering
    \includegraphics[width=1\linewidth]{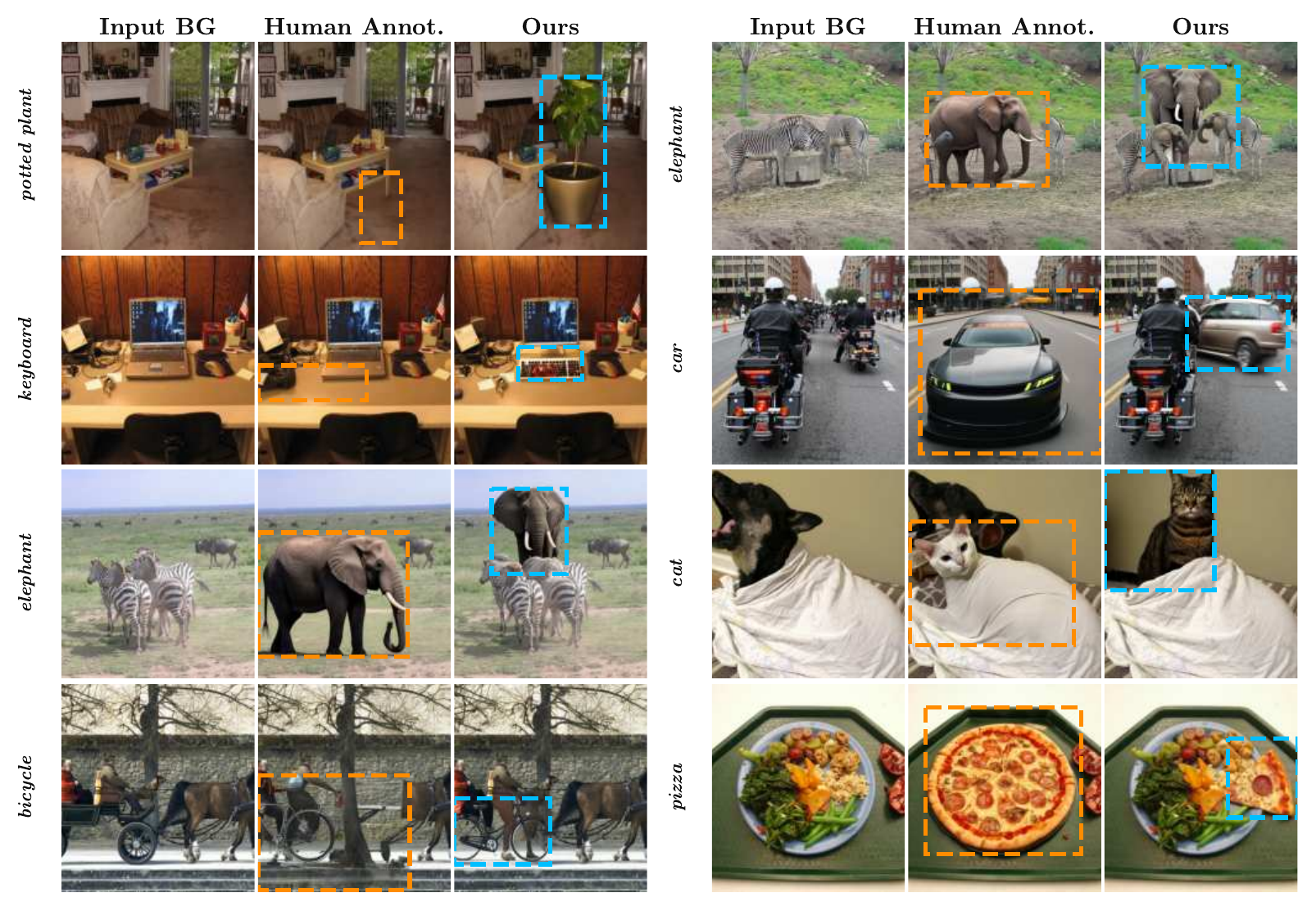}
    \caption{
\textbf{Image inpainting with object placement priors.}
We evaluate three object placement approaches on the \textsc{OPA} backgrounds.
\emph{Human Annot.} uses ground-truths from OPA.
\emph{Random BBox} samples a uniformly random placement region, serving as a
controlled baseline.
\emph{Ours} uses the top-ranked box of our dataset.
Dashed boxes show the inpainting region
(\textcolor[HTML]{00BFFF}{HiddenObject annotation (ours)},
 \textcolor[HTML]{FF8C00}{OPA annotation},
 \textcolor[HTML]{CC44CC}{Random}).
}
    \label{fig:suppmat_small}
\end{figure}

\section{Image Editing}
\label{sec:supp_image_editing}

\subsection{Quantitative Evaluation via ImgEdit-Judge}

We extend the image-editing and placement results from Fig.~\ref{fig:qualitative_imgedit}, comparing
insertions guided by human annotations, random boxes, and our spatial priors.
For our method, we use the top ImageReward-ranked annotation whose IoU with the
largest Grounding-DINO-detected background object is below 50\%, preventing its
replacement. Figure~\ref{fig:suppmat_small} shows inpainting results;
Figs.~\ref{fig:placement_a}--\ref{fig:placement_d} provide additional examples.

\subsection{Human Evaluation}
\label{sec:supp_human_eval}

\begin{figure}[t]
\centering
\includegraphics[width=\linewidth]{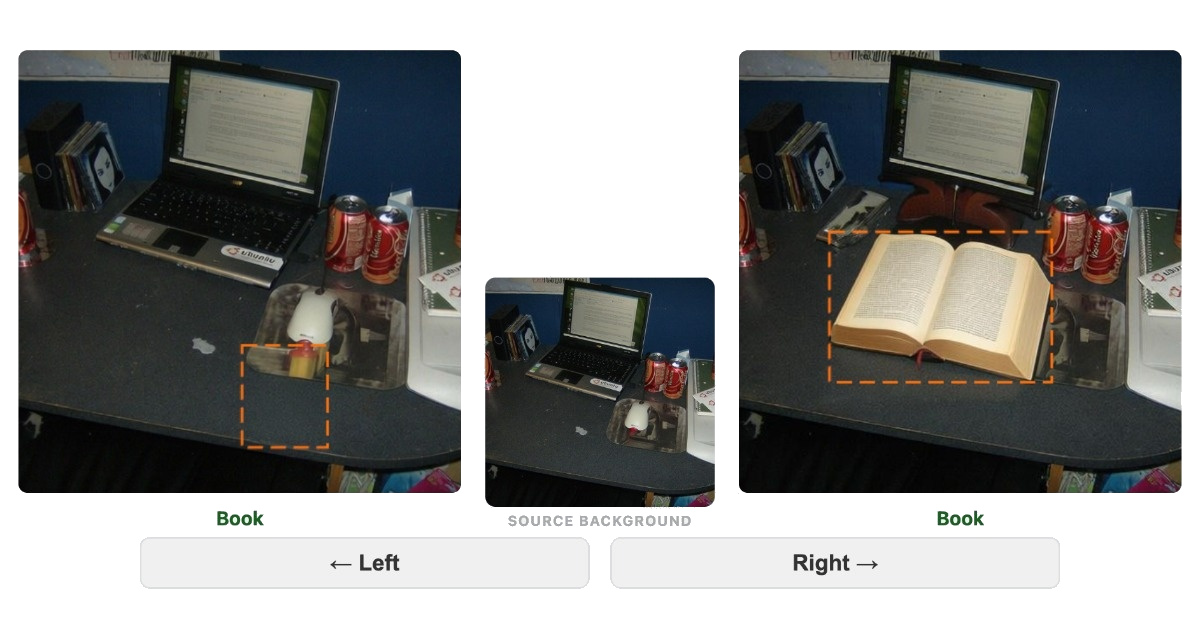}
\caption{
\textbf{Human evaluation interface for inpainting preference assessment (Study A).}
Participants compare two edits of the same scene and object category with
different insertion locations. They select the more realistic and coherent
image based on placement, scale, and overall realism.
}
\label{fig:human_a}
\end{figure}

\begin{figure}[t]
\centering
\includegraphics[width=\linewidth]{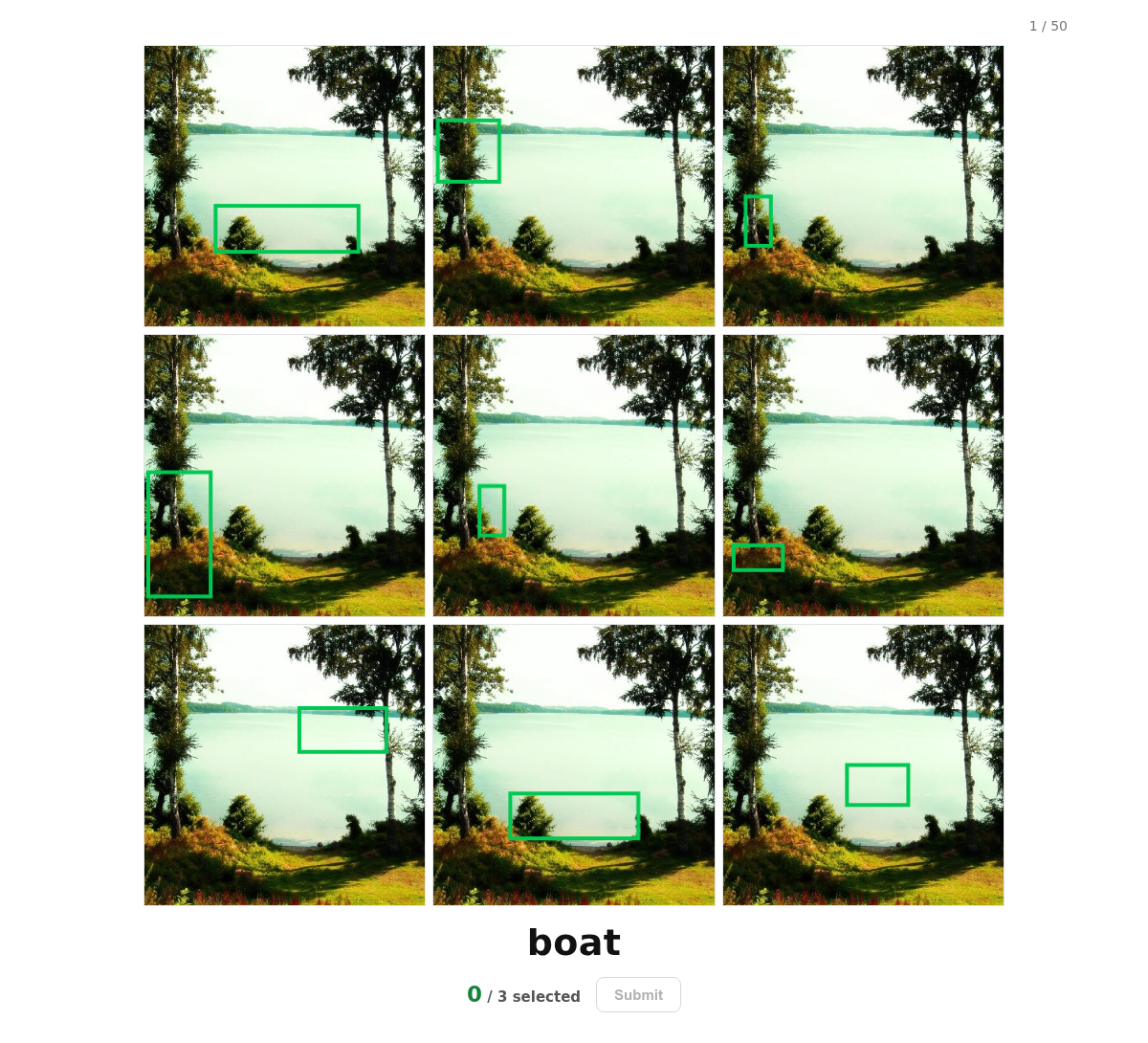}
\caption{
\textbf{Human evaluation interface for bounding-box plausibility assessment (Study B).}
Participants select the three most plausible boxes from a $3\times3$ grid for
one scene--object pair, considering position, scale, and aspect ratio.
}
\label{fig:human_b}
\end{figure}

\begin{figure}[t]
\centering
\includegraphics[width=\columnwidth]{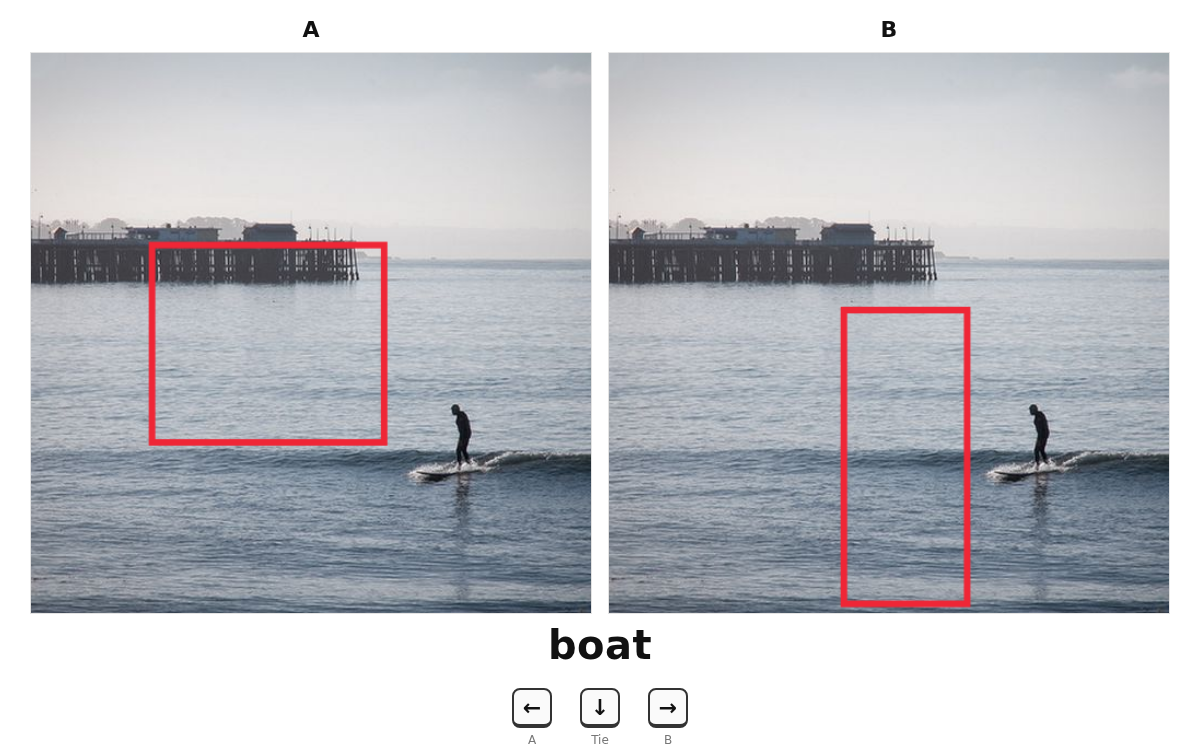}
\caption{
\textbf{Human evaluation interface for placement preference assessment (Study C).}
Participants compare two boxes on the same clean background and select A, B,
or a tie based on placement and scale.
}
\label{fig:human_c}
\end{figure}

Figures~\ref{fig:human_a}--\ref{fig:human_c} show the study interfaces. We
conducted all evaluations anonymously with standardized instructions and asked
participants to make immediate visual judgments.

\paragraph{Study A: Inpainting Preference.}
Participants compared two generated images with the same background and object
category but different insertion locations. They selected the more realistic
image based on placement, scale, visual consistency, and scene coherence. The
object category was shown for context. The study included 76 \textsc{OPA}
pairs.

\paragraph{Study B: Bounding-Box Plausibility.}
Participants selected exactly three plausible boxes from a $3\times3$ grid for
each scene--object pair. They considered only position, scale, and aspect ratio,
isolating placement from synthesis quality. The study used 50 unique
\textsc{HiddenObjects} scene--object pairs.

\paragraph{Study C: Placement Preference.}
In a blinded A/B study, participants compared two boxes on the same background
for the same target class, without inpainted objects. They selected the box with
the more plausible position and scale or marked a tie. The study used all eligible 79 unique
\textsc{OPA} scene--object pairs and 20 unique randomly sampled \textsc{HiddenObjects} scenes.

\section{Spatial Prior Visualization}
\label{sec:supp_prior_viz}

Building upon the aggregated spatial prior distributions introduced in Fig.~\ref{fig:cherry_picked_dataset} of the main paper, we present a comprehensive set of randomly sampled spatial prior visualizations across  25 distinct object categories. These qualitative examples are detailed in Figs.~\ref{fig:pa}--\ref{fig:pc}. To construct the heatmaps, ranking scores are first normalized across candidate boxes using a softmax, after which the resulting weights are accumulated over the pixels contained within each bounding box to form a spatial density map. The aggregated map is then min--max normalized and blended with the input image to produce the final visualization overlay. We aggregate only those bounding box annotations that achieve a verification detector confidence greater than $0.4$ and a positive preference ranker score (${r_{ij}} > 0$). This robust filtering step ensures that the resulting spatial priors strictly reflect valid, highly preferred, and visually coherent object placements. 

\section{Qualitative Analysis of Pipeline Components}
  \label{sec:supp_component_qualitative}

  We complement the quantitative component analysis in
Sec.~\ref{sec:key_componenet_selection} with qualitative comparisons of the
  inpainter and ranker. These examples illustrate characteristic differences
  between components; our final selections remain based on the aggregate OPA
  results in Tab.~\ref{tab:component_selection}.

\begin{figure}[t]
      \centering
      \includegraphics[width=\columnwidth]{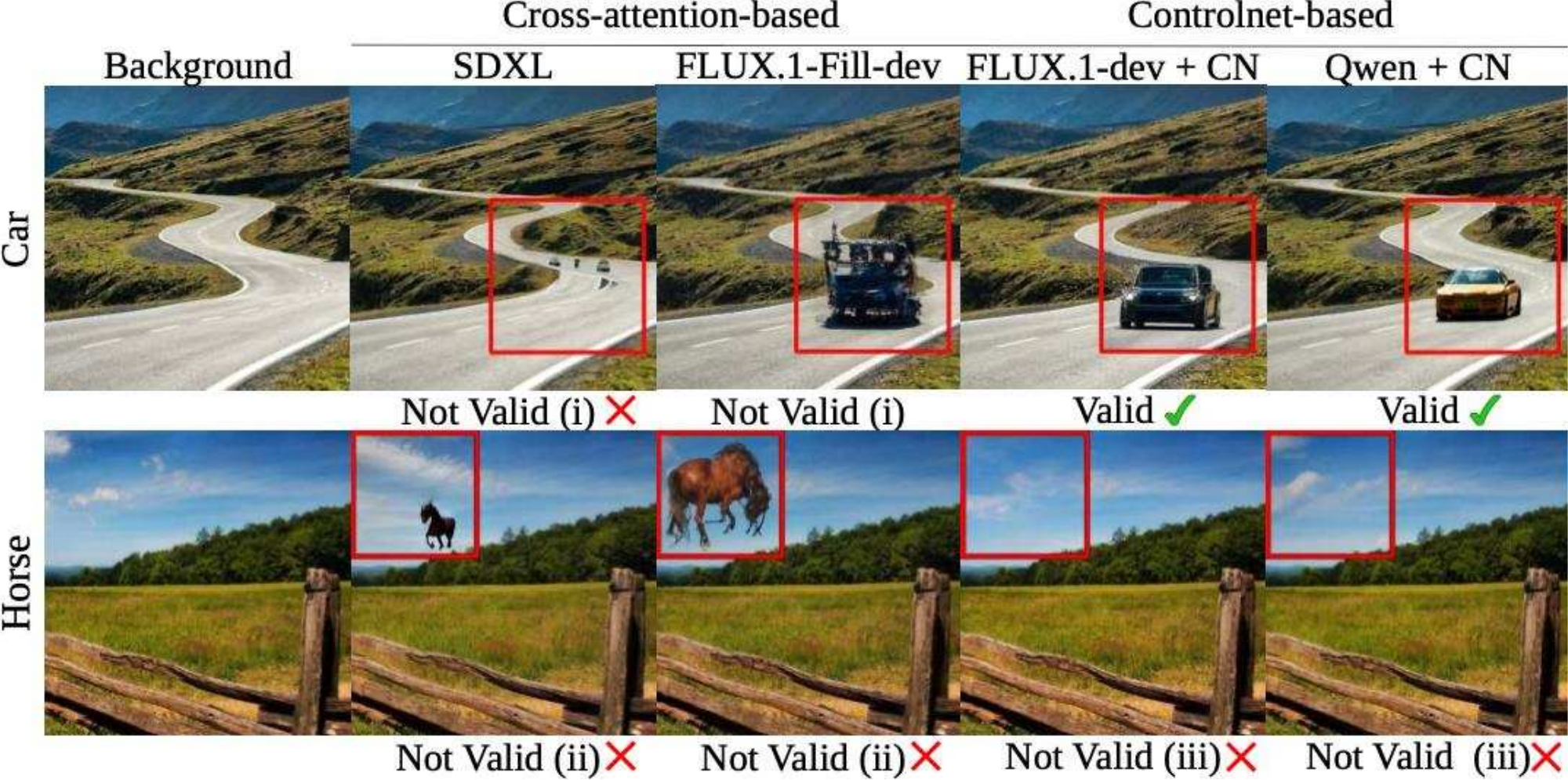}
      \caption{
      \textbf{Qualitative inpainter comparison.}
      Each row fixes the background, target class, and candidate region.
      Standard inpainters may produce artifacts (i) or disregard scene geometry
      (ii). ControlNet-based variants better respect structural constraints,
      producing valid insertions or refusing incompatible placements (iii).
      }
      \label{fig:inpainter_comparison}
  \end{figure}

  \mypar{Inpainter Selection.}
  We compare inpainters using the same background, target class, and candidate
  box. As shown in Fig.~\ref{fig:inpainter_comparison}, standard
  prompt-conditioned inpainters such as SDXL~\cite{podell2023sdxl} and
  FLUX~\cite{batifol2025flux} tend to synthesize the requested object even when
  the candidate location conflicts with the scene geometry. This can produce
  distorted objects or semantically implausible placements, such as a horse
  floating in the sky. ControlNet-based variants preserve the surrounding
  structure more strongly: they produce better-grounded objects for feasible
  locations and may refuse to generate an object at an incompatible location.
  In our pipeline, such refusals are subsequently rejected by the object
  detector and are preferable to retaining a visually implausible insertion.
  Together with its highest F1 score in Tab.~\ref{tab:component_selection},
  these observations motivate our use of Qwen+CN~\cite{wu2025qwen,qwencn}.

\begin{figure}[t]
      \centering
      \includegraphics[width=\columnwidth]{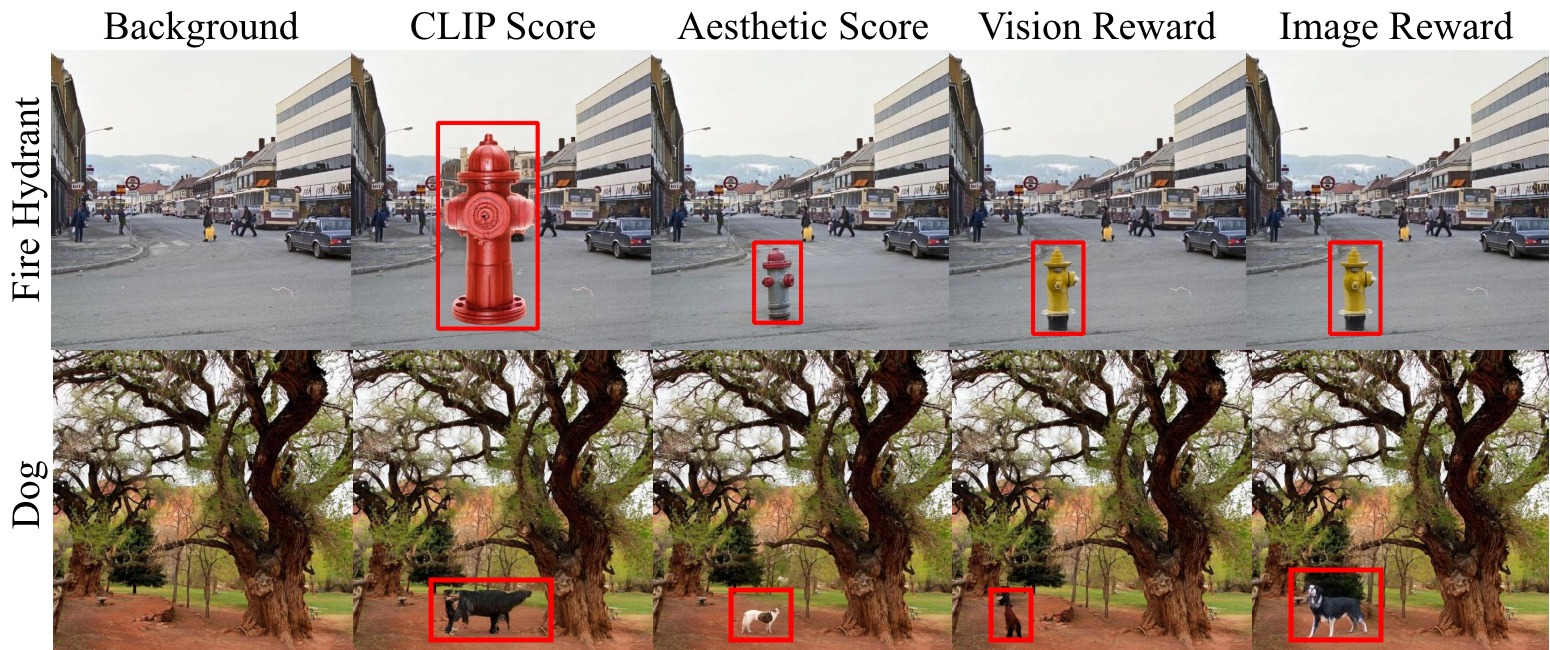}
      \caption{
      \textbf{Qualitative ranker comparison.}
      We show the highest-scoring insertion selected by each ranker from the
      same candidate pool. CLIP Score can prioritize semantic visibility,
      whereas reward-based rankers better account for contextual realism,
      scale, and grounding.
      }
      \label{fig:ranker_comparison}
  \end{figure}

  \mypar{Ranker Selection.}
  We next fix the inpainter and candidate set and vary only the function used to
  rank the surviving insertions. Figure~\ref{fig:ranker_comparison} shows the
  highest-scoring candidate selected by each ranker. The examples reveal that
  CLIP Score~\cite{hessel2021clipscore} can prioritize semantic visibility over
  global composition, as illustrated by the oversized fire hydrant. Aesthetic
  Score~\cite{aestheticpredictor2024} favors visually clean outputs but does not
  explicitly assess whether an object has a context-appropriate scale or
  location. Reward models account more directly for the realism of the complete
  edited image. In the shown examples, ImageReward~\cite{xu2023imagereward}
  selects objects with more plausible grounding and scale. This qualitative
  behavior is consistent with its highest AP in
  Tab.~\ref{tab:component_selection}, motivating its use as our ranker.

\section{Placement Model}
\label{sec:supp_detr_impl}

As discussed in the main paper, our distillation model for learning object placement from \textsc{HiddenObjects} is a simple variant of DETR. Here, we provide details of the architecture and training setup.

\begin{figure}[!t]
    \centering
    \includegraphics[width=1\linewidth]{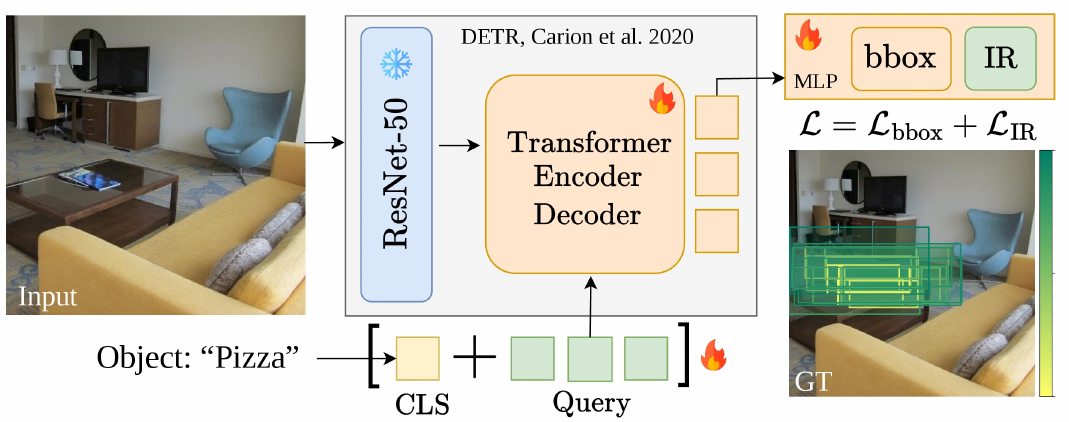}
\caption{\textbf{Placement Distillation Model.}
Given a background image and a target object class, the model predicts plausible insertion locations using a DETR-style architecture. The transformer decoder operates on a fixed set of learned queries that are conditioned on the target class embedding. Each query predicts a candidate bounding box with its plausibility score. During inference, predicted boxes are ranked by their plausibility scores to obtain the most likely placement locations for the target object.}
    \label{fig:architecture}
\end{figure}

\mypar{Architecture.}
We formulate object placement as a multi-class, closed-set prediction task. Our architecture is illustrated in Fig.~\ref{fig:architecture}. The input is a center-cropped RGB image of size $512\times512$. We extract multi-scale visual features using a frozen ResNet-50 backbone and process them with a DETR-style Transformer encoder-decoder with sinusoidal positional encoding. The encoder consists of 6 layers with hidden dimension $256$, 8 attention heads, and feedforward dimension $2048$.

The target category (e.g., ``apple'', ``cat'') is mapped to a learnable class embedding $c \in \mathbb{R}^{B\times256}$. This embedding is replicated across $|Q|=50$ placement queries and added to learnable query offsets, producing the final $\mathbb{R}^{B\times50\times256}$ input to the decoder.

We apply two prediction heads to the $|Q|$ decoded queries. A 3-layer MLP bounding-box head with hidden dimension $256$ and ReLU activations predicts four normalized values corresponding to the top-left box coordinate, width, and height. In parallel, a 2-layer MLP plausibility head regresses the ImageReward score. During inference, the model produces 50 candidate placements per image, ranked by the sigmoid of their predicted plausibility scores.

\mypar{Training.}
We train using Hungarian matching with a regression loss combining L1 and GIoU, weighted by the normalized reward scores. We use the top 20 bounding boxes from our spatial prior with the highest reward scores for supervision. We optimize using AdamW with a learning rate of $1\mathrm{e}{-4}$, weight decay of $1\mathrm{e}{-4}$, and gradient clipping at a norm of $0.1$. We train with a batch size of 128 for up to 200 epochs, using early stopping with a patience of 15 epochs.

\mypar{Loss.} We first perform bipartite matching between predictions and ground-truth boxes using the Hungarian algorithm, minimizing a cost function $\mathcal{C}$ that combines $L_1$ distance and Generalized Intersection over Union (GIoU). For matched pairs, the spatial loss is defined as:
\begin{equation}
    \mathcal{L}_{\text{bbox}} = \omega \left[ \lambda_{L1} \|b - \hat{b}\|_1 + \lambda_{giou} (1 - \text{GIoU}(b, \hat{b})) \right]
\end{equation}
where $\omega = 0.5 + 0.5R$ is a weighting factor derived from the image-reward score $R$. The reward head is trained to regress the maximum IoU of each query relative to the ground-truth set. Let $\hat{p}_j$ be the logit for query $j$. The plausibility loss is:
\begin{equation}
    \mathcal{L}_{\text{plaus}} = \frac{1}{N} \sum_{j=1}^{N} \| \sigma(\hat{p}_j) - \max_k \text{IoU}(\hat{b}_j, b_k) \|^2
\end{equation}
Our final objective is $\mathcal{L} = \mathcal{L}_{\text{bbox}} + 0.5\mathcal{L}_{\text{plaus}}$, with $\lambda_{L1}=5.0$ and $\lambda_{giou}=2.0$.

\section{Inpainting Speedup}
\label{sec:supp_speedup}

The main computational bottleneck in computing spatial priors is the diffusion-based inpainting stage. For each image, we evaluate 1,004 candidate boxes, only a small number of which correspond to valid insertions (i.e., confidence greater than 0.4). As discussed in the main paper, to detect this, we introduce an early-stopping detection process based on divergence in denoising performance in the early denoising steps. Below, we detail our early stopping heuristic derivation.

\begin{figure}[!t]
    \centering
    \includegraphics[width=\linewidth]{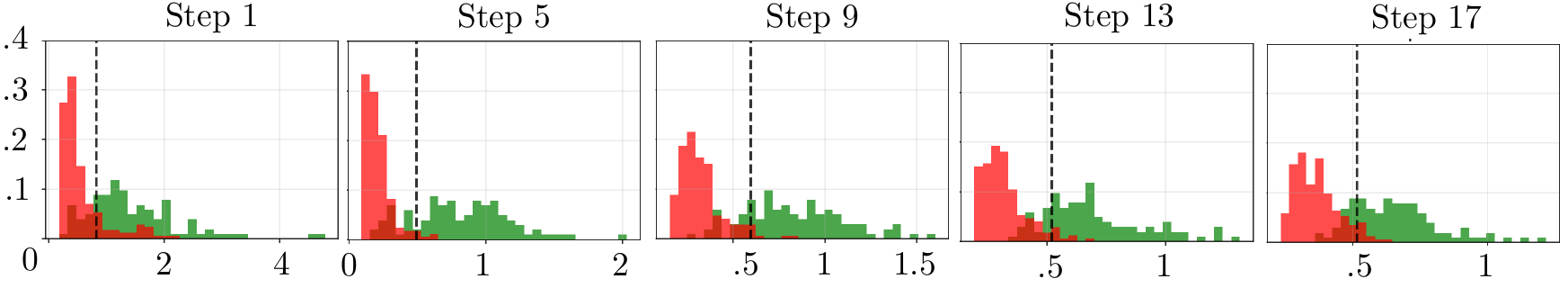}
    \includegraphics[width=\linewidth]{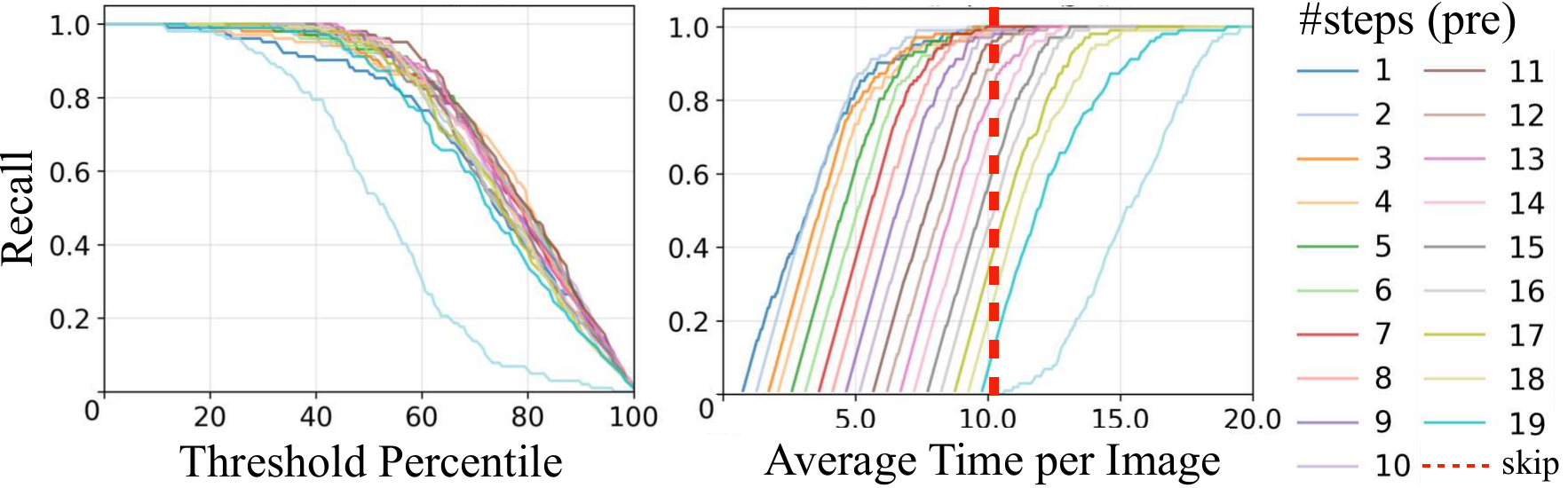}
\caption{\textbf{Speedup Analysis.} \textbf{Top}: Distribution of $\Delta(\mathbf{x}_t, \varnothing)$ for successful (green) and failed (red) inpaintings at selected denoising steps, computed on 15060 inpaintings. The dashed line marks the most separating threshold at each step. \textbf{Bottom:} Recall versus threshold percentile (left) and versus  average time per image (right), with one curve per number of denoising steps run before the exit check. Checking after the second denoising step is sufficient to reach 80\% recall at a 2.2$\times$ speedup. The dashed \textcolor{red}{red} line (\emph{skip}) denotes the cost of fully inpainting all boxes, $N\cdot \tau_{20}$.}
   \label{fig:speedup}
\end{figure}

\mypar{Criterion.} Given a diffusion model $\varepsilon_{\theta}$ that denoises a noisy image $\mathbf{x}_t$ at step $t$ by computing the noise $\varepsilon_{\theta}(\mathbf{x}_t, c)$ for condition $c$. An image is inpainted by taking the difference between the conditional branch, e.g., $c=\text{``cat''}$ and the unconditional branch $c=\varnothing$. Our observation is simple: we measure how much these two noise estimates diverge throughout denoising:

\begin{equation}
    \Delta(\mathbf{x}_t, \varnothing) = \|\varepsilon_{\theta}(\mathbf{x}_t, c) - \varepsilon_{\theta}(\mathbf{x}_t, \varnothing)\|_{1}.
\end{equation}
In the top part of Fig.~\ref{fig:speedup}, we compute this distribution per denoising step for $15$ random images and observe how the distribution of  $\Delta(\mathbf{x}_t, \varnothing)$ changes across timestep. We notice that there is a fine separation between successful and unsuccessful generations, where successful ones exhibit much higher values (mean $1.47$) than failed ones (mean $0.38$) at the second step. While in step-5 generations are slightly better separated, to speedup computation we have to trade separability with efficiency.

\mypar{Threshold computation.} If $N$ is the number of boxes and $s(r_{\lambda})$ is the  fraction of all boxes passing threshold $\lambda$ for a given recall $r$ and $\tau_{t}$ is the computational time required until step $t$, the total computational time per recall $\tau(r_{\lambda})$:
\begin{equation}
 \tau(r_{\lambda}) = N(\tau_t + s(r_{\lambda})\tau_{20}),
\end{equation}
\noindent where the  first term corresponds to the check, while the second to the full inpainting. We use $t=2$, as the second denoising step gives the best trade-off. As shown in Fig.~\ref{fig:speedup}, we find that a calibrated threshold $\lambda = 0.7148$ achieves 80.4\% recall after two denoising steps, with only $s(r_{\lambda})=0.34$ fraction of bounding box locations being passed to full inpainting, leading to a 2.2x speedup.

\begin{figure}[!t]
    \centering
    \includegraphics[width=\linewidth]{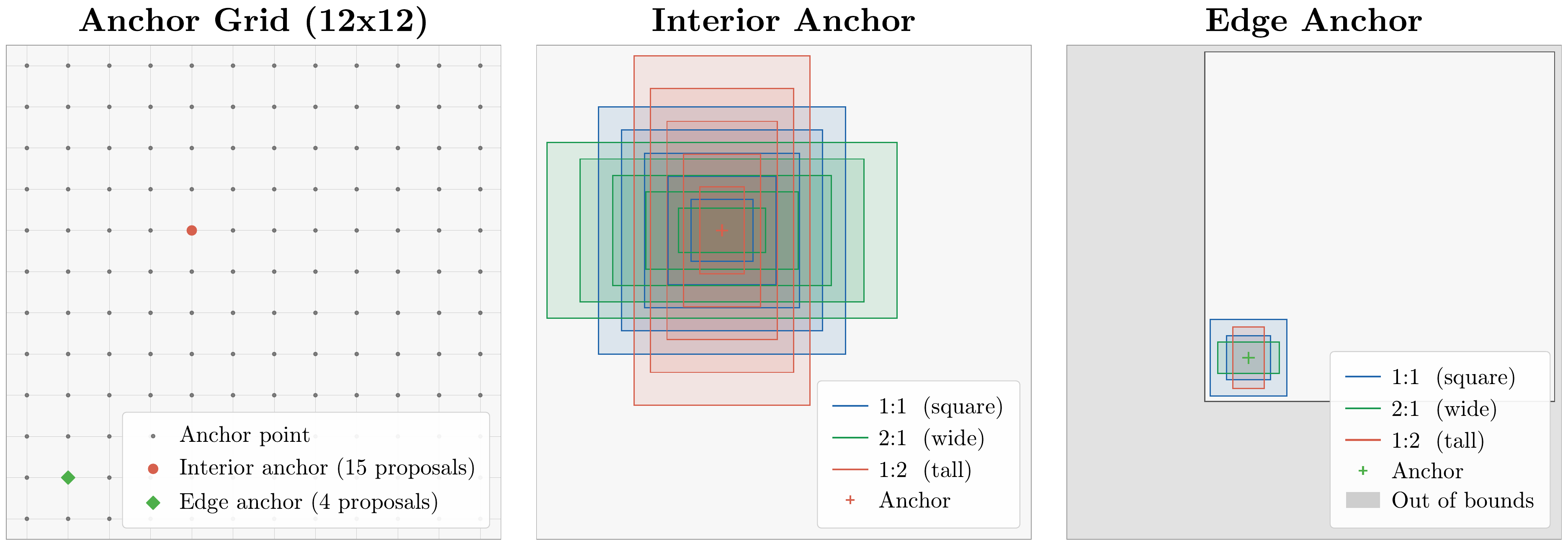}
    \caption{\textbf{Bounding Box Proposals.} \textbf{Left}: The $12 \times 12$ anchor grid. \textbf{Middle}: Full coverage set for an interior anchor. \textbf{Right}: Pruned set for a corner anchor due to boundary constraints.}
    \label{fig:bbox_viz}
\end{figure}

\section{Bounding box Proposals}
\label{supmat:bbox}
Bounding box proposals for inpainting are produced using a sliding window approach, explained in Fig.~\ref{fig:bbox_viz}. Given an image of size $S \times S$ and a target number of proposals $N$, we first derive a square grid of $G \times G$ anchor points, where $G = \lfloor \sqrt{\lfloor N/3 \rfloor} \rfloor$. These anchors are uniformly spaced with a half-stride offset to provide symmetric padding from the image boundaries.

For each anchor, we generate candidate boxes by combining three aspect ratios $\mathcal{R} = \{1:1, 2:1, 1:2\}$ with five linearly spaced scales $s \in [s_{\min}, s_{\max}]$. To ensure area-equivalence across ratios, the width and height are computed as $w = s \sqrt{r_w/r_h}$ and $h = s \sqrt{r_h/r_w}$. A candidate box $\mathbf{b}$ is added to the final proposal set $\mathcal{B}$ only if it is strictly contained within the image manifold $[0, S] \times [0, S]$.

We set $S=512$, $N=435$, and a scale range of $[64, 256]$. While the unconstrained maximum number of boxes is 2,160 ($144 \text{ anchors} \times 15$), the boundary constraint prunes edge-adjacent candidates to consider only fully visible objects. This deterministic configuration yields a final set of $|\mathcal{B}| = 1,004$ bounding boxes.

\section{Zero-Shot VLM Placement Protocol}
\label{sec:supp_vlm_placement_protocol}

\paragraph{Setup.}
We evaluate Qwen2.5-VL-72B/7B, InternVL3-8B, and LLaVA-OneVision-7B on
545 background--class queries. Each model receives only the clean background
and target class; ground truth is used only for evaluation. Images are
center-cropped and resized to $512\times512$. We use each checkpoint's official
chat template with the following shared prompt:

\begin{vlmprompt}{Zero-shot ranked object-placement prompt}
\ttfamily\scriptsize\raggedright\sloppy

\promptrole{SYSTEM}\quad You are an object-placement expert. Follow the coordinate convention and ranking instructions exactly. Return only the requested JSON object.

\promptrole{USER}\quad \promptvar{IMAGE}

You are given a clean $512\times512$ pixel background image and the target object class ``\promptvar{OBJECT CLASS}.'' Predict exactly 5 distinct bounding boxes at spatially different plausible locations where one instance of the target object could be placed naturally and realistically. Do not repeat a box or location; all five box centers must differ.

Rank predictions from most to least plausible using ranks 1--5. Assign each a real-valued confidence in $[0,1]$ representing how natural the placement is. Confidence must be non-increasing with rank. Produce five alternatives even when some have low confidence, and do not use rank numbers as confidence scores.

Use absolute pixel corner coordinates with origin at the upper-left. Each bbox has fields \texttt{x1, y1, x2, y2}, where $(x1,y1)$ is the upper-left corner and $(x2,y2)$ the lower-right. Verify $0 \leq x1 < x2 \leq 512$ and $0 \leq y1 < y2 \leq 512$ for every box. \texttt{x2} and \texttt{y2} are corner coordinates, not width and height. Each box should cover the full expected extent of the inserted object, not a point or support surface.

Return only:\
\char123"predictions": [\
\quad\char123"rank": 1, "bbox": \char123\
\qquad"x1": 0, "y1": 0, "x2": 1, "y2": 1\char125,\
\qquad"confidence": 0.0\char125,\
\quad...,\
\quad\char123"rank": 5, "bbox": \char123\
\qquad"x1": 0, "y1": 0, "x2": 1, "y2": 1\char125,\
\qquad"confidence": 0.0\char125\
]\char125

\end{vlmprompt}

\paragraph{Inference.}
Schema-constrained decoding requires five ranked boxes with confidences in
$[0,1]$. 

\paragraph{Evaluation.}
For standard class-level mAP@0.50, detections are pooled by class, ranked by the
reported confidence, and matched one-to-one to ground-truth boxes. We compute
PASCAL VOC 11-point AP per class and average over the 28 classes. IoU@1/5 is the
best overlap with any retained target among the first one/five boxes, and
IoU50@1/5 is the corresponding success rate. These scores measure agreement
with pipeline-retained placements, not all human-plausible placements.

\section{Background - Foreground Assignment}
\label{app:bgfg}
To produce general enough coverage of object placement, we define 10  macro-categories of objects: person, vehicles, outdoor items, animals, sport items, kitchen items, food, furniture, technology, and miscellaneous. Then, we manually aggregate 50 COCO classes per macro-category. Finally, we employ a manually defined object-to-scene taxonomy between the 50 foreground object classes from COCO and the 126 distinct background classes from Places365 to minimize non-meaningful object-background placement requests. The full assignment is detailed in Tab.~\ref{tab:bg_fg_assignment}.

\clearpage
\begin{table*}[!ht]
\centering
\caption{Assignment between COCO foreground and Places365 background classes.}
\label{tab:bg_fg_assignment}

\begin{adjustbox}{max width=\textwidth, max height=0.9\textheight}
\scriptsize
\setlength{\tabcolsep}{3pt}
\begin{tabular}{llp{10cm}}
\toprule
\textbf{Category} & \textbf{COCO Foreground} & \textbf{Places365 Background} \\
\midrule

\rowcolor{clr_person} person & person & alley, amphitheater, apartment building, atrium, attic, ballroom, desert sand, downtown, market, martial arts gym, plaza, shopping mall, street, temple, wheat field \\
\rowcolor{clr_person} person & umbrella & attic, balcony exterior, balcony interior, boardwalk, campus, dining hall, patio, street \\
\rowcolor{clr_person} person & suitcase & airport terminal, atrium, attic, bus station, car interior, hotel room, motel \\
\rowcolor{clr_person} person & handbag & atrium, attic, car interior, coffee shop, department store, restaurant, shopping mall \\
\rowcolor{clr_person} person & backpack & airport terminal, atrium, attic, campus, car interior, forest path, park \\
\midrule

\rowcolor{clr_vehicle} vehicles & bicycle & attic, campus, downtown, park, promenade, residential neighborhood \\
\rowcolor{clr_vehicle} vehicles & car & auto factory, desert road, garage, gas station, highway, parking lot, street \\
\rowcolor{clr_vehicle} vehicles & motorcycle & desert road, garage, gas station, highway, parking lot, street \\
\rowcolor{clr_vehicle} vehicles & airplane & airfield, hangar indoor, hangar outdoor, runway, sky \\
\rowcolor{clr_vehicle} vehicles & boat & boat deck, boathouse, harbor, lagoon, lake, ocean, river \\
\midrule

\rowcolor{clr_outdoor} outdoor & traffic light & downtown, driveway, promenade, runway, street \\
\rowcolor{clr_outdoor} outdoor & bench & courtyard, park, patio, promenade, residential neighborhood \\
\rowcolor{clr_outdoor} outdoor & fire hydrant & downtown, park, residential neighborhood, runway, street \\
\rowcolor{clr_outdoor} outdoor & stop sign & courtyard, crosswalk, downtown, residential neighborhood, street \\
\rowcolor{clr_outdoor} outdoor & parking meter & downtown, parking lot, promenade, residential neighborhood, street \\
\midrule

\rowcolor{clr_animal} animals & dog & alley, botanical garden, chalet, field wild, kennel, park, residential neighborhood, veterinarians office, yard \\
\rowcolor{clr_animal} animals & cat & alley, balcony, bedroom, botanical garden, chalet, house, living room, veterinarians office, yard \\
\rowcolor{clr_animal} animals & elephant & bamboo forest, desert sand, field wild, forest path, rainforest \\
\rowcolor{clr_animal} animals & horse & cliff, farm, field wild, kennel, pasture, stable, tundra \\
\rowcolor{clr_animal} animals & cow & barn, corn field, farm, field wild, kennel, pasture \\
\midrule

\rowcolor{clr_sports} sports & kite & athletic field, beach, chalet, field wild, mountain path, park \\
\rowcolor{clr_sports} sports & skateboard & campus, chalet, garage, park, playground, promenade, street \\
\rowcolor{clr_sports} sports & tennis racket & apartment building, athletic field, chalet, gymnasium, park, playground \\
\rowcolor{clr_sports} sports & surfboard & beach, boat deck, coast, garage, pier, wave \\
\rowcolor{clr_sports} sports & sports ball & athletic field, baseball field, basketball court, gymnasium, park, soccer field \\
\midrule

\rowcolor{clr_kitchen} kitchen & bottle & bar, cafeteria, coffee shop, dining room, kitchen, restaurant, restaurant patio \\
\rowcolor{clr_kitchen} kitchen & cup & cafeteria, coffee shop, dining room, fastfood restaurant, kitchen, restaurant \\
\rowcolor{clr_kitchen} kitchen & wine glass & banquet hall, bar, coffee shop, dining room, restaurant, restaurant patio \\
\rowcolor{clr_kitchen} kitchen & spoon & bar, cafeteria, coffee shop, dining room, kitchen, restaurant \\
\rowcolor{clr_kitchen} kitchen & fork & bar, cafeteria, coffee shop, dining room, kitchen, restaurant \\
\midrule

\rowcolor{clr_food} food & sandwich & cafeteria, coffee shop, fastfood restaurant, picnic area, restaurant, restaurant patio \\
\rowcolor{clr_food} food & apple & dining room, kitchen, market, picnic area, vegetable garden \\
\rowcolor{clr_food} food & cake & bakery shop, banquet hall, cafeteria, dining room, picnic area, restaurant \\
\rowcolor{clr_food} food & pizza & dining room, fastfood restaurant, picnic area, pizzeria, restaurant \\
\rowcolor{clr_food} food & orange & dining room, kitchen, market, picnic area, vegetable garden \\
\midrule

\rowcolor{clr_furni} furniture & potted plant & attic, balcony interior, formal garden, living room, office, patio, zen garden \\
\rowcolor{clr_furni} furniture & chair & attic, basement, beach house, bedroom, classroom, dining room, living room, office \\
\rowcolor{clr_furni} furniture & couch & attic, basement, beach house, hotel room, house, living room, reception, television room \\
\rowcolor{clr_furni} furniture & dining table & banquet hall, dining room, kitchen, restaurant, restaurant patio \\
\rowcolor{clr_furni} furniture & bed & attic, bedchamber, bedroom, dorm room, hotel room \\
\midrule

\rowcolor{clr_tech} technology & tv & attic, basement, bedroom, home theater, hotel room, living room, television room \\
\rowcolor{clr_tech} technology & laptop & attic, basement, bedroom, dorm room, home office, library, office \\
\rowcolor{clr_tech} technology & cell phone & basement, bedroom, dorm room, home office, office, reception \\
\rowcolor{clr_tech} technology & keyboard & basement, bedroom, computer room, dorm room, home office, office \\
\rowcolor{clr_tech} technology & microwave & basement, cafeteria, dining room, kitchen, office, restaurant kitchen \\
\midrule

\rowcolor{clr_misc} misc. & book & attic, basement, bookstore, classroom, dorm room, library, living room, office \\
\rowcolor{clr_misc} misc. & toothbrush & bathroom \\
\rowcolor{clr_misc} misc. & clock & attic, basement, berth, classroom, house, living room, office, reception \\
\rowcolor{clr_misc} misc. & scissors & art studio, artists loft, basement, classroom, fabric store, home office, office \\
\rowcolor{clr_misc} misc. & vase & archive, attic, basement, bedroom, house, living room, reception \\
\bottomrule
\end{tabular}
\end{adjustbox}
\end{table*}

\newcommand{\imgdir}{img/appendix/}
\foreach \l/\f/\t in {
a/selection_page_01.jpg/{airplane, apple, bench, bicycle, boat, book, bottle, cake.},
b/selection_page_02.jpg/{car, cat, chair, cow, cup, dog, elephant, fire hydrant.},
c/selection_page_03.jpg/{horse, keyboard, laptop, motorcycle, person, pizza, potted plant, sandwich.}
}{%
  \begin{figure*}[!ht]
    \centering \includegraphics[width=0.95\textwidth,height=0.90\textheight,keepaspectratio]{\imgdir\f}
    \caption{\textbf{Spatial Prior}. Object placements are computed for the following  classes: \t}
    \label{fig:p\l}
  \end{figure*}
}

\renewcommand{\imgdir}{img/appendix/}

\foreach \l/\f/\t in {
a/fig_suppmat_v2_1.pdf/{airplane, apple, backpack, bed, bench, bicycle, boat, book.},
b/fig_suppmat_v2_2.pdf/{bottle, cake, car, cat, cell phone, chair, clock, couch.},
c/fig_suppmat_v2_3.pdf/{cow, cup, dining table, dog, elephant, fire hydrant, fork, handbag.},
d/fig_suppmat_v2_4.pdf/{horse, keyboard, kite, laptop, microwave, motorcycle, orange, parking meter.}
}{%
  \begin{figure*}[!ht]
    \centering
    \includegraphics[width=0.95\textwidth,height=0.90\textheight,keepaspectratio]{\imgdir\f}
    \caption{
\textbf{Inpainting with Object Placement Priors.}
We evaluate three object placement approaches on the \textsc{OPA} backgrounds.
\emph{Human Annot.} uses ground-truths from OPA.
\emph{Random BBox} samples a uniformly random placement region, serving as a
controlled baseline.
\emph{Ours} uses the top-ranked box of our dataset.
Dashed boxes show the inpainting region
(\textcolor[HTML]{00BFFF}{HiddenObject annotation (ours)},
 \textcolor[HTML]{FF8C00}{OPA annotation},
 \textcolor[HTML]{CC44CC}{Random}).
}
    \label{fig:placement_\l} 
  \end{figure*}
}

\end{document}